%% file: neurips.tex
\documentclass{article}

\usepackage[eandd,final]{neurips_2026}

\usepackage[utf8]{inputenc} % allow utf-8 input
\usepackage[T1]{fontenc}    % use 8-bit T1 fonts
\usepackage{url}            % simple URL typesetting
\usepackage{booktabs}       % professional-quality tables
\usepackage{amsfonts}       % blackboard math symbols
\usepackage{nicefrac}       % compact symbols for 1/2, etc.
\usepackage{microtype}      % microtypography
\usepackage{xcolor}         % colors
\usepackage{xspace}
\usepackage{graphicx}
\usepackage{float}
\usepackage{multirow} 
\usepackage{wrapfig}
\usepackage{amssymb}
\usepackage{amsmath}
\usepackage{enumitem}
\usepackage{hyperref}     
\hypersetup{
    colorlinks=true,
    linkcolor=blue,
    urlcolor=blue,
    citecolor=blue,
    filecolor=blue,
}

\definecolor{blue}{HTML}{B86046}

\definecolor{codegreen}{rgb}{0,0.6,0}
\definecolor{codegray}{rgb}{0.5,0.5,0.5}
\definecolor{codepurple}{rgb}{0.58,0,0.82}
\definecolor{backcolour}{rgb}{0.95,0.95,0.92}
\definecolor{bg}{rgb}{0.95,0.95,0.95}
\definecolor{mygreen}{rgb}{0,0.6,0}

\newcommand{\benchmarklogo}{\raisebox{-1pt}{\includegraphics[width=.95em]{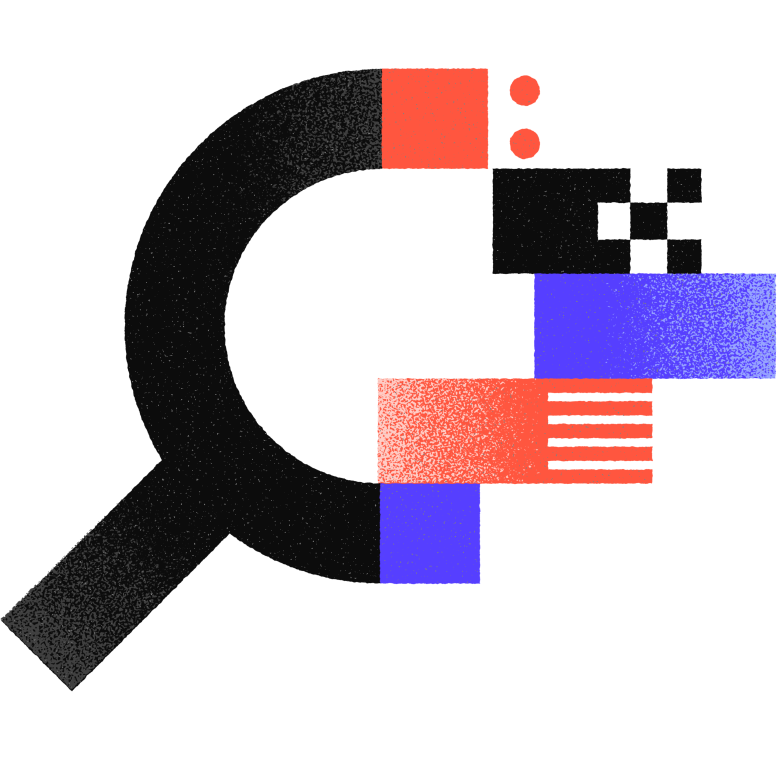}}}
\newcommand{\benchmarktext} {\textsc{CodeReviewBench}\xspace}
\newcommand{\benchmark}{\mbox{\benchmarktext\benchmarklogo}\xspace}

\title{\benchmark: An Automated Benchmark for Evaluating the Software Factory}

\author{%
    Ashley Zhang\textsuperscript{1,*} \quad
  Narmeen Fatimah Oozeer\textsuperscript{1} \quad
  Aleksandr Zverianskii\textsuperscript{1} \quad
  Daniel Ng\textsuperscript{1} \\
  Tasana Pejovic\textsuperscript{1} \quad
  Ant\'ia Garc\'ia\textsuperscript{1} \quad
  Jacob Clyne\textsuperscript{1} \quad
  Ravpreet Setia\textsuperscript{2} \\
  Akshay Utture\textsuperscript{3} \quad
  Etan Jacob Ginsberg\textsuperscript{1} \quad
  Shriyash Kaustubh Upadhyay\textsuperscript{1} \quad
  Fazl Barez\textsuperscript{1,4} \\
  \\
  \textsuperscript{1}Martian \quad
  \textsuperscript{2}Greptile \quad
  \textsuperscript{3}Augment \quad
  \textsuperscript{4}TSG Lab, University of Oxford \\
  \textsuperscript{*}Correspondence: \texttt{ashley@withmartian.com}
}

\begin{document}

\maketitle

\begin{abstract}
\input{abstract}
\end{abstract}

\section{Introduction}

\input{introduction}

\section*{Acknowledgements and Disclosures}

We thank Erfan Al-Hossami and David Loker (CodeRabbit), Itamar Friedman and Dedy Kredo (Qodo), Paul Senglé-Ferrière (Cubic), and Guy Eisenkot and Nimrod Kor (Baz) for feedback on benchmark design and interpretation.  We also thank Jack Smith for developing MIRA and supporting the experiments reported in this paper, and Sam Jung for contributions to benchmark data quality.

\paragraph{Funding.}
This work was conducted at and funded by Martian.

\paragraph{Competing interests.}
Most authors are affiliated with Martian, which maintains \benchmarktext. Ravpreet Setia is affiliated with Greptile, and Akshay Utture is affiliated with Augment. Both companies develop code-review tools evaluated in the benchmark, and the offline gold set builds in part on benchmark data they previously released. We independently audited and revised those data before evaluation. The authors declare no other competing interests.

\bibliographystyle{plainnat}
\bibliography{references}

%%%%%%%%%%%%%%%%%%%%%%%%%%%%%%%%%%%%%%%%%%%%%%%%%%%%%%%%%%%%

\appendix

\input{appendix.tex}

%%%%%%%%%%%%%%%%%%%%%%%%%%%%%%%%%%%%%%%%%%%%%%%%%%%%%%%%%%%%

\end{document}

%% file: abstract.tex
As AI agents ship more code with minimal human oversight, code review becomes the critical quality-control checkpoint — but one without a reliable, objective verifier: review quality is nuanced and ultimately a judgment call by the PR author. Existing benchmarks are structurally misaligned with this: annotators are third-party reviewers, like the bots they evaluate, not the PR authors who actually decide what to fix. They also rely on small, static issue sets that require full re-evaluation as tools update, limiting how frequently progress can be tracked. We introduce \textbf{\benchmark}, built on a fundamentally different signal: \textit{revealed developer preferences} — if a developer acts on a bot's suggestion, we treat it as useful; if they ignore it, we do not. This signal is mined daily from the public GitHub event stream, yielding 500K+ judged PRs across 18 tools, and refreshes continuously as tools evolve. We validate the developer preference through human annotation and against an independent offline benchmark: the two agree on precision rankings and independently detect the same product updates within days of vendor releases. Together, these form the benchmark.

%% file: introduction.tex
\label{sec:intro}

Benchmarks are our proxies for AI capability; when they break, so does our ability to measure progress. Code review sits at the center of two measurement problems at once. First, it is the quality-control layer of the automated software factory, but is increasingly being bypassed: in 2025, pull requests (PR) on GitHub surged 23\% and commits jumped 25\%, yet review comments grew only 0.35\%~\citep{github2025octoverse}, and over half of bot-reviewed pull requests already see zero human activity after the bot's review (\S\ref{sec:quality}). Second, and less visibly, code review is the reward signal that trains code generators: Reinforcement Learning (RL)-based training requires a judgment about whether generated code is correct, and that judgment requires code review. A flawed benchmark for review tools is therefore not merely a leaderboard problem; it corrupts the reward function for the broader generation pipeline.

Evaluating AI code review is harder than evaluating code generation. Code generation admits an objective verifier: a patch either passes the hidden unit tests or it does not. A review comment has no such test. Its usefulness is in part a private judgment by the pull-request author, depending on correctness, whether the author already knew the issue, project priorities, and the interpretive frame of the team~\citep{bacchelli2013expectations}. Even code generation, with its objective verifier, has struggled: SWE-bench Verified~\citep{openai2024swebenchverified}, the field's standard for 18 months, was retired in February 2026 after frontier models were found to reproduce gold patches verbatim from training data~\citep{openai2026whyweretire}. Static benchmarks on public code are structurally vulnerable to leakage, and the smaller the dataset, the faster it saturates. Code review is harder still: the SWE-rebench response of refreshing the task stream while keeping the verifier does not apply when no verifier exists.

Prior work approximates a verifier by curating gold sets of expert-labeled issues~\citep{greptile2025, augment2025}, but this embeds a structural misalignment no amount of annotation effort can resolve: annotators are third-party reviewers, like the bots they evaluate, not the PR authors who actually decide what to fix. Beyond this, gold sets are incomplete in both directions: the same 50 pull requests curated multiple independent times each find real issues the previous pass missed with no evidence of convergence, while 6 golden issues were caught by zero of 20 tracked tools and 41 clusters of concerns flagged by $\geq$3 tools are absent from the gold set entirely (App.~\ref{app:gold-set-analysis}). Sets of 50--137 issues also lack statistical power and cannot be refreshed at the pace tools update, and the result is predictable: every vendor that publishes a benchmark wins its own~\citep{deepsource2026blog}. These are not incidental flaws but structural ones, which is why we develop a new signal rather than a better-curated set.

We therefore change the signal from what annotators assert to what developers do. If a developer pushes a commit addressing a bot suggestion, it was useful; if they ignore it, it was not (though some useful suggestions are not acted on; \S\ref{sec:limitations}). The public GitHub event stream records these responses across millions of repositories, and gaming them requires changing real developer behavior at scale, making the benchmark naturally resistant to Goodhart's Law~\citep{goodhart1984}.

\textbf{We make the following contributions}: 

\begin{itemize}[nosep]
    \item \textbf{A scalable, automated pipeline} that mines AI code-review activity from the public GitHub event stream daily, links each tool's suggestions to subsequent developer actions, and scores them via an LLM judge validated across three model families (Fleiss's $\kappa = 0.86$). Because the pipeline runs continuously, the benchmark naturally resists Goodharting and detects product-level capability shifts as they happen.

    \item \textbf{One of the largest online datasets for code review,} pairing ${\sim}$500K+ judged pull requests across 18 tools (refreshed daily) with a 50-PR offline gold set of 136 expert-labeled issues (extending \citealp{greptile2025, augment2025})---over three orders of magnitude larger than existing code-review benchmarks and $20\times$ the scale of SWE-rebench. Each online PR pairs tool suggestions with developer responses, providing both evaluation signal and fine-tuning data for open-source code-review models.

    \item \textbf{Convergent validation from both benchmarks.} Behavioral data at scale is noisy; quality filtering and multi-stage validation are needed to recover a trustworthy signal. We validate internally (robustness, replicability, fairness) and externally (against the offline gold set, human expert annotations, and vendor changelogs), establishing that rankings reflect tool capability rather than measurement artifacts.

    \item \textbf{A configurable public leaderboard} stratified by PR size, severity, language, and domain---dimensions inferred by an LLM labeling pass not directly observable from repository metadata---enabling engineering teams to evaluate tools against their specific workflow rather than a single aggregate ranking.
\end{itemize}

% Our central methodological claim is that revealed developer preferences are a viable measurement primitive for code review, complementary to expert curation. Each captures a different facet of tool quality, and we provide convergent evidence from both that the same underlying capabilities are being measured (\S\ref{sec:offline}).

All code, prompts, the offline gold set, an online data sample, and the live dashboard are open source.\footnote{\url{https://github.com/withmartian/code-review-benchmark}} 
Section~\ref{sec:methodology} describes the pipeline; Section~\ref{sec:quality} establishes the behavioral signal through internal validation; Section~\ref{sec:offline} shows where the two benchmarks converge and diverge; Section~\ref{sec:findings} presents findings; Section~\ref{sec:applications} describes broader use cases of \benchmarktext beyond evaluation.

\section{Methodology}
\label{sec:methodology}
\input{methodology}

%% file: methodology.tex
To measure code-review quality from revealed preferences, we need to observe what a bot suggested and whether the developer acted on it. The pipeline (Figure~\ref{fig:pipeline}) automates, drawing from fresh real-world PRs daily so that scores track live tool behavior and cannot be gamed by optimizing against a fixed dataset. It transforms raw GitHub events into scored PR records through five stages:

\begin{figure}[t]
\centering
\includegraphics[width=0.87\textwidth]{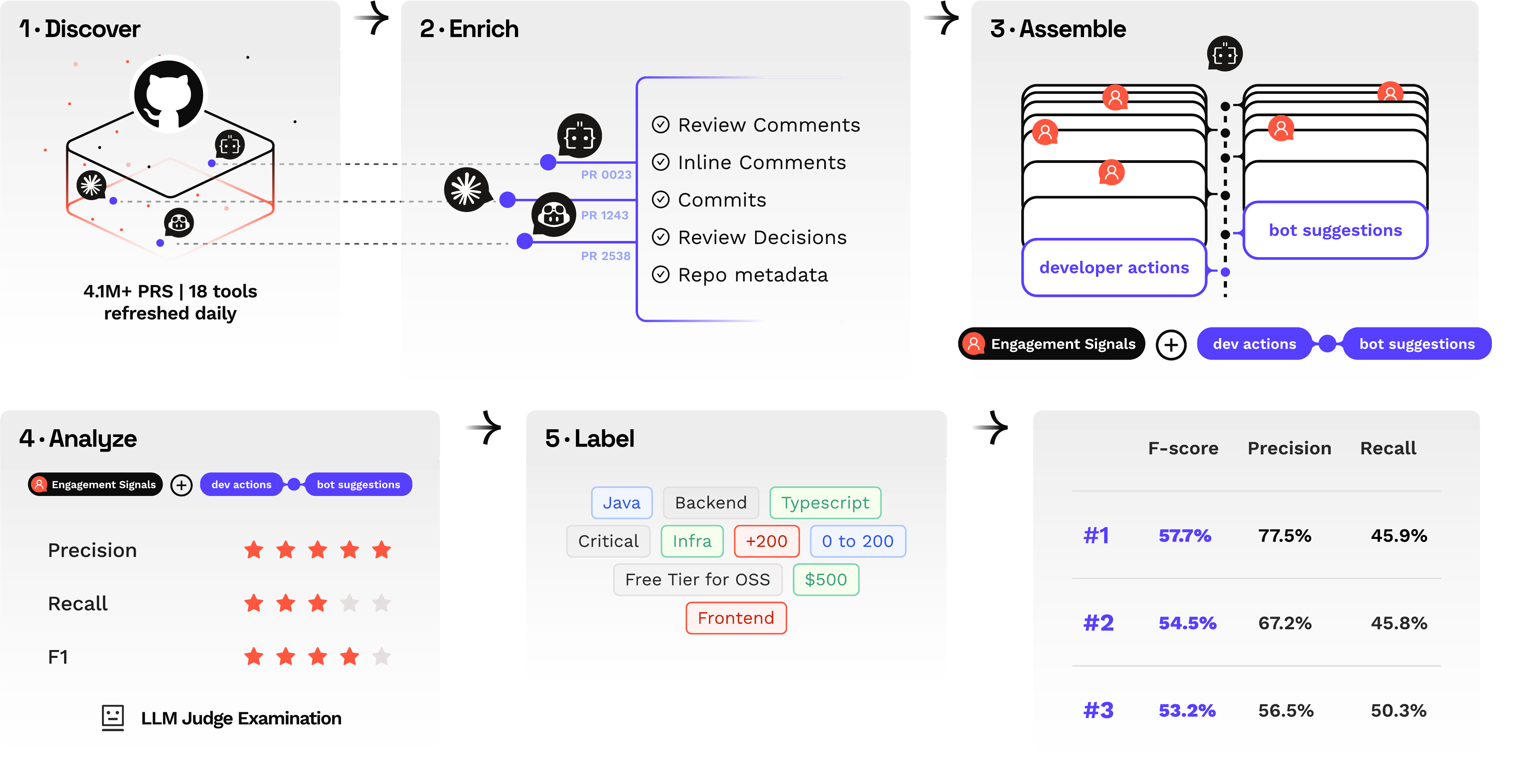}
\caption{The Code Review Bench pipeline: (1)~Discover PRs via BigQuery, (2)~Enrich via GitHub API, (3)~Assemble bot--developer timelines, (4)~Analyze with LLM judge, (5)~Label for stratification.}
\label{fig:pipeline}
\end{figure}

\begin{description}
    \item[Discover.] A daily BigQuery scan over GitHub Archive~\citep{githubarchive} identifies PRs on which any of 25 tracked AI review bots has commented, sampling up to 500~PRs per bot per day.
    \item[Enrich.] The GitHub REST API retrieves full context for each candidate PR: diff, review and inline comments, commits before and after review, review decisions, and repository metadata.
    \item[Assemble.] Per-PR timelines are constructed pairing each bot suggestion with the developer actions that follow it (commits, comments, and review decisions). Human engagement signals (reviewer count, comment count, back-and-forth review rounds, post-review commits) are computed per PR for use in quality filtering (\S\ref{sec:quality}).
    \item[Analyze.] An LLM judge decides whether each bot suggestion was addressed by a subsequent developer action and assigns a score (\S~\ref{sec:judge}). Only merged PRs are analyzed: unmerged PRs lack reliable developer-action signal.
    \item[Label.] A second LLM pass classifies each PR by language, domain, PR type, and severity for stratified analysis exposed through the configurable leaderboard.
\end{description}
Collected data accumulates in a PostgreSQL database: the pipeline runs daily, processing new pull requests as they appear.

\paragraph{Data Volume.}

The pipeline has discovered {4.1M+} merged pull requests reviewed by 25~AI code-review tools, of which {$\sim$900k+} are fully enriched across $\sim${190K} repositories and {200+} languages (Table~\ref{tab:bot-coverage} in Appendix~\ref{app:coverage}). Of these, {$\sim$500k+} have been analyzed by the LLM-judge and {160K+} filtered by quality yield a scorable precision--recall pair. We report results for the 18 tools with sufficient volume for stable scoring (\S\ref{sec:quality}).

Tool coverage mirrors deployment volume: Copilot appears in $\sim$143K enriched PRs across $\sim$70K repositories; the smallest tools contribute $\sim$1K each. Per-repository concentration varies (2.0 PRs/repo for Copilot to 21.5 for Graphite), which our quality filtering and concentration cap address (\S\ref{sec:quality}).

\paragraph{LLM judge.}
\label{sec:judge}

The assembled timeline splits commits at the bot's review point: pre-review commits form the context the bot saw; post-review commits and comments constitute the developer's response. We use GPT-5 Nano as an LLM judge in a three-step structured-output prompt:

\begin{enumerate}
    \item \textbf{Extract bot suggestions.} The judge reads the pre-review diff and the bot's review comments, extracting each distinct actionable suggestion with a category label (bug, security, performance, style, \ldots) and a severity label (low, medium, high, critical).
    \item \textbf{Extract developer actions.} The judge reads the post-review diffs, comments, and review decisions, identifying each distinct issue that was concretely addressed.
    \item \textbf{Match suggestions to actions.} For each bot suggestion, the judge determines whether it identifies substantially the same concern as any extracted developer action, allowing partial overlap: a suggestion that caught part of a real problem counts as a true positive. Unmatched suggestions are false positives; unmatched actions are false negatives.
\end{enumerate}

We separate extraction from matching so that each step has a focused task, reducing compound errors. All matching is within a single PR; cross-PR causal tracing is left to future work (\S\ref{sec:limitations}). Full prompt text is in Appendix~\ref{app:prompts}.

\textbf{Judge validation.}
We validate the judge along two axes, namely \emph{model stability} and \emph{methodology robustness}, using a stratified sample of 180~PRs from the online database (all 15 leaderboard tools, balanced across precision terciles). Three judge families re-run extraction and matching independently, and an alternative Comment Action Rate judge bypasses extraction entirely to test methodology sensitivity (See details in \S\ref{sec:internal-validation}). Full agreement matrices are in Appendix~\ref{app:judge-validation}.

\paragraph{Metrics.}
\label{sec:metrics}
We score tools with precision, recall, and F$_\beta$.

\textbf{What each metric measures.}
Per-PR \emph{precision} is the fraction of a tool's suggestions matched to a developer action and answers ``when this tool speaks, is it useful?'' Per-PR \emph{recall} is the fraction of developer-addressed issues matched to a suggestion, which answers ``of the issues that mattered, how many did the tool catch?'' We macro-average both across all scored PRs (weighting each PR equally regardless of suggestion count).

The harmonic mean F$_\beta$ combines precision and recall with a configurable weight: $\beta > 1$ favors recall, $\beta < 1$ favors precision, and $\beta = 1$ (F$_1$) weights them equally. We report F$_1$ as the default within-benchmark metric for two reasons: (1)~it balances the extremes (pure precision rewards silence while pure recall rewards verbosity) and (2)~this balance is harder to game, since optimizing one side often penalizes the other. The dashboard exposes adjustable~$\beta$ so practitioners can shift toward their preference. Because the precision--recall gap varies substantially across tools, the choice of~$\beta$ materially affects rankings (\S\ref{sec:findings}; Appendix~\ref{app:beta}).

\textbf{When do metrics transfer across evaluation settings?}
A metric is transferable if it produces consistent tool rankings across independent evaluation setups. We expect this to vary by metric:

\begin{itemize}
    \item \textbf{Precision} is most likely to transfer because both benchmarks penalize low-value suggestions, whether developers ignore them online or experts omit them from the offline gold set.
    \item \textbf{Recall} should not fully transfer: what counts as a ``miss'' differs structurally. Online, a valid issue may not count because the developer defers it due to project priorities, style, or timing. Offline, a miss is an expert-curated issue the tool does not catch; experts review more comprehensively but may lack the PR author's context and priorities. These are different questions about coverage, and neither replaces the other.
    \item \textbf{F$_1$} inherits recall's divergence. Cross-benchmark agreement should rise as $\beta$ decreases and precision receives more weight. Full F$_1$ agreement would require developers to act on every valid suggestion and the gold set to cover every issue they value, neither of which holds in practice.
\end{itemize}

\noindent We test these predictions in \S\ref{sec:divergence}: precision rankings transfer across benchmarks, while recall and F$_1$ do not reliably transfer.

\section{Establishing the Behavioral Signal}
\label{sec:quality}
\input{establishing_behavioral_signal}

%% file: establishing_behavioral_signal.tex
The online benchmark scores tools by what developers do, not what annotators say. Whether that produces a meaningful signal depends on several things being true at once: the LLM judge has to read the data correctly, the data quality filters we apply have to recover signal rather than create it, and the rankings shouldn't simply reflect the kinds of repositories each tool happens to be deployed in. Mining 500K+ PRs produces abundant data, but much of it is noisy, so we proceed in three stages: quality filtering to remove uninformative PRs (\S\ref{sec:noise}); internal validation testing \emph{robustness} (invariance to judge model and filter choice), \emph{replicability} (stability across time splits and resampling), and \emph{fairness} (rankings not driven by deployment context); then external validation against the independent offline benchmark and vendor changelogs (\S\ref{sec:offline}).

\paragraph{The noise problem.}
\label{sec:noise}
Of 518K bot-reviewed merged PRs, more than half show zero human activity after the bot's review (Figure~\ref{fig:engagement-raw} in Appendix~\ref{app:engagement-raw}). Without a later commit or comment, there is no observable action to match suggestions against: silence could mean the review was unhelpful or that the developer never looked. Without filtering, rankings would reflect repository habits rather than tool behavior. We expose six quality filters (Table~\ref{tab:filters}); the default combination retains 41\% of the data. The question is whether these filters recover a signal or create one. 

Beyond noise removal, the filters also provide structural resistance to gaming: the concentration cap prevents any single author--repo pair from dominating a tool's score, and the engagement filter ensures that only PRs where a developer visibly responded contribute. This adds prevention of inflation or deflation of scores by deploying in repositories where reviews are systematically ignored. We discuss gaming vectors and additional countermeasures in \S\ref{sec:limitations} and Appendix~\ref{app:goodhart}.

\begin{table}[h]
\centering
\label{tab:filters}
\small
\begin{tabular}{llr}
\toprule
\textbf{Filter} & \textbf{Effect} & \textbf{Retained} \\
\midrule
Base            & All merged PRs with a tracked bot review        & 100\%  \\
Bot authors     & Exclude PRs opened by bot accounts              & 88.6\% \\
Conc.\ cap     & Sample $\leq$50 PRs per author--repo--bot triple  & 83.8\% \\
Engagement      & Require PRs with $\geq$1 human comment or commit after bot review & 41.3\% \\
\midrule
Team repos      & Require PRs with $\geq$2 human contributors in repo     & 20.2\% \\
Human reviewer  & Require PRs with $\geq$1 non-author human reviewer       &  9.3\% \\
\bottomrule
\end{tabular}
\caption{Quality filters applied cumulatively; each row adds one filter to all prior rows. Data retention is measured across the Feb--Apr 2026 window as a percentage of the unfiltered Base set. The first four filters form the paper's default; the last two are optional strict layers.}
\end{table}

\paragraph{Internal validation.}
\label{sec:internal-validation}
We test the signal's robustness, replicability, and fairness.

\paragraph{Instrument robustness.}
A robust measurement is invariant to reasonable changes in the setup: swapping the judge model or adjusting the data quality filter configuration should not materially change tool rankings. Three judges from different model families, run independently on the same data, agree closely on which suggestions were addressed. An alternative methodology that skips the extraction step and matches comments directly to diffs reaches nearly identical tool-level scores. Rankings also hold across six filter configurations (permissive to strict) and shift only marginally when contributor/reviewer thresholds are swept. Ranking stability confirms the filters remove noise rather than manufacture a result; the filters remain necessary because they tighten confidence intervals and exclude PRs that lack developer action.

\paragraph{Replicability.}
A replicable measurement produces the same rankings when repeated on different samples drawn from the same window. Chronological split-halves agree closely; 1{,}000 random day-partitions give the same answer; rankings hold at multiple window sizes. A day-block bootstrap places the typical score CI within ${\sim}$1\,pp, which is tight enough to distinguish most pairs of tools.

\paragraph{Fairness.}
A fair measurement does not let confounds from non-random deployment drive the rankings. Our setup is observational: teams pick tools for reasons that correlate with the PRs those tools review. A tool deployed mostly in solo-developer repositories faces an easier scoring environment than one deployed in active multi-reviewer repositories. We use Double Machine Learning~\citep{chernozhukov2018dml} to estimate each tool's effect on per-PR precision (chosen because precision is well-defined per PR and is the most portable metric across evaluation contexts, \S\ref{sec:offline}) after controlling for nine confounders (diff size, language, repo activity, etc.). Debiasing attenuates raw differences but does not reorder the rankings suggesting that the selection bias is not dominant.  Additional sensitivity checks (engagement--score correlation, rank stability across the full $F_\beta$ range, language/domain stratification) are in Appendices~\ref{app:beta},\ref{app:engagement-corr},\ref{app:stratification}; Table~\ref{tab:validation-summary} in the appendix summarizes all internal tests.

\paragraph{What internal validation cannot establish.}
\label{sec:internal-limits}

The tests in \S\ref{sec:internal-validation} show the pipeline produces a stable, precise, judge-robust signal. They do not show that it measures tool quality rather than something else. If developers systematically act on easy suggestions rather than useful ones, no amount of internal consistency would catch it. That requires comparing to a measurement built on entirely different ground, which is what \S\ref{sec:offline} and \S\ref{sec:human-validation} do.

\paragraph{Human validation.}
\label{sec:human-validation}

The online benchmark measures \emph{developer-aligned usefulness}: whether the PR author acts on a review suggestion. This is not absolute review quality, which is not directly observable. Developers have richer project context but act selectively based on their priorities; expert annotators assess comments more comprehensively but have less project context. We use two human annotation tasks as convergent evidence rather than treating either source as definitive ground truth.

\paragraph{Task~1: Does the judge match suggestions correctly? ($n = 125$).} Experts independently repeated the matching decision using each bot suggestion and the post-review developer actions extracted for that PR. Experts agreed with the automated judge on 85.6\% of decisions (Cohen's $\kappa = 0.71$). This result validates suggestion--action matching, conditional on the extracted action list, not the completeness of the preceding extraction stages.

\paragraph{Task~2: Does developer action track comment quality? ($n = 150$).} Experts assessed bot comments against the full PR diff without seeing the judge output or developer response. Annotators judged 60.7\% of comments as identifying real issues, including issues fixed in a later commit. Across 11 tools, human-assessed accuracy correlates with automated precision at Spearman $\rho = 0.688$ ($p = 0.019$). This provides independent evidence that developer action preserves a meaningful quality signal.

\noindent Together, these tasks show that the judge's matches generally agree with experts and that automated precision tracks expert-assessed comment quality. Detailed study design, statistics, weaknesses, and error analysis are provided in Appendix~\ref{app:human-validation}.

\section{Cross-Benchmark Validation}
\label{sec:offline}
\input{cross_primitive_convergence}

%% file: cross_primitive_convergence.tex
The two benchmarks have nearly opposite strengths. The online benchmark captures revealed behavior at scale and refreshes daily, but its signal is uncontrolled. The offline benchmark scores every tool on the same 50 PRs against expert-curated golden issues, providing controlled comparison but limited to a fixed set re-curated at most a few times a year. Together they support a workflow neither supports alone: the online benchmark detects when something changes; the offline benchmark verifies the change under controlled conditions. A full table of structural contrasts is in Appendix~\ref{app:contrasts}.

\paragraph{Rank agreement.}
\label{sec:rank-agreement}
Across the 11 tools present in both leaderboards, precision rankings agree at $\rho = 0.69$ ($p = 0.019$; Table~\ref{tab:cross-benchmark}). Two measurements built on different ground truths identify the same tools as the precise ones. The P/R ratio agrees at $\rho = 0.60$, indicating each tool's position on the precision--recall trade-off is preserved. F$_1$ does not agree; we return to why in~\S\ref{sec:divergence}.

With $n = 11$, a natural concern is that agreement could be driven by a few outliers. A leave-one-out analysis shows this is not the case: removing any single tool keeps $\rho$ positive, and bootstrap 95\% CIs are $[+0.12, +0.94]$. The one partial outlier is Qodo (online rank 10, offline rank 5), explained by a version transition during the measurement window where 66\% of its online data comes from a deprecated version scoring 13\,pp lower (\S\ref{sec:changepoints}; App.~\ref{app:changepoints}). Excluding it raises $\rho$ to $+0.78$.

Convergence under different operationalizations is the standard argument for construct validity~\citep{construct_validity_ref}: when two measurements built on different foundations agree on what they jointly measure, the construct is real.

\paragraph{Detecting and verifying real-world changes.}
\label{sec:changepoints} 
When a vendor ships an update that changes review quality, both benchmarks should detect the shift independently, confirming they track the same underlying capability rather than correlated artifacts.

We apply the PELT changepoint detector~\citep{killick2012} to each tool's daily online F$_1$ and find eight significant level shifts across six tools (Figure~\ref{fig:changepoints}, App.~\ref{app:changepoints}). The shifts are staggered rather than synchronized ($p_{\text{sync}} \approx 1.0$), providing evidence against a single benchmark-wide judge or pipeline change. Five of the eight align with public vendor changelogs, in several cases within a day of the announced release. For Qodo, an internal version marker independently reproduces the pre- and post-changepoint means exactly (App.~\ref{app:changepoints}).

Five of these changepoints have paired offline before/after runs. All five agree on the direction of the F$_1$ change ($p = 0.031$); four agree within 3\,pp on magnitude. The one outlier had its before-run three weeks before the changepoint, capturing a longer improvement arc (Figure~\ref{fig:tracking-changes}, App.~\ref{app:tracking-changes}).

This is the trigger-and-verify pattern: the online benchmark identifies when a tool's behavior changes, and a targeted offline re-run confirms the change under controlled conditions.

\paragraph{Where the benchmarks diverge.}
\label{sec:divergence} 
Disagreements between the benchmarks are also informative, and confirm the transferability expectations laid out in \S\ref{sec:metrics}. Three kinds appear in the data.

\textbf{Regime-driven.} Online, the ground truth is the PR author: whether the developer who wrote the code acts on a suggestion. Offline, it is a third-party annotator, structurally closer to the bot's own perspective (both are outside reviewers). PR authors fix selectively based on their priorities and context (online P/R $= 1.53$); annotators label comprehensively (offline P/R $= 0.78$). As predicted in \S\ref{sec:metrics}, precision and the P/R ratio transfer across benchmarks ($\rho = 0.69$ and $0.60$) while F$_1$ does not ($\rho = -0.05$). Sweeping $\beta$ from recall-weighted toward precision-weighted confirms the pattern: cross-benchmark agreement rises monotonically as precision's weight increases (Appendix~\ref{app:beta}).

\textbf{Coverage-driven.} The offline gold set is incomplete in both directions. Clustering tool comments on the 50-PR set by textual similarity yields 41 clusters where $\geq$3 independent tool families flag the same concern absent from the gold set (roughly 30\% of its size) (App.~\ref{app:gold-set-analysis}). In the other direction, six golden issues were caught by zero of 20 tools. This incompleteness biases offline scores: precision is underestimated (real finds scored as false positives) and recall is overestimated (denominator too small). The online benchmark, scoring against developer behavior instead, does not share this bias, so comparing the two locates where the gold set has gaps.
\textbf{Temporal.} The offline benchmark is a point-in-time snapshot; the online benchmark is a rolling average. For tools that improved during the measurement window, the online average understates current performance while a recent offline run reflects only the latest state. This is by design: the online benchmark detects when a tool changes, and a targeted offline re-run confirms what changed.
\paragraph{Why this pattern matters.}
\label{sec:workflow-implication}

Gold sets go stale, and continuous re-curation is impractical at the scale needed. Neither benchmark alone is sufficient: the online benchmark cannot self-validate, and the offline benchmark is too small and too static. Together they form a closed loop: the online benchmark flags where to look; the offline benchmark confirms what happened. Each covers the other's blind spots.

Finally, each divergence type points to a concrete improvement path: regime-driven gaps motivate calibrating the gold set to developer priorities rather than annotator priors; coverage gaps identify where new golden issues should be added (or removed); and temporal gaps set the re-curation cadence needed to keep offline scores meaningful.

\section{Findings}
\input{findings}

%% file: findings.tex
\label{sec:findings}
All results in this section use the default quality filters (exclude bot-authored PRs, require human engagement, concentration cap of 50~PRs per repo-author pair) over the Feb~16--Apr~20, 2026 window unless stated otherwise. We include the 15~tools with $\geq$2{,}000 total PRs in the time frame and $\geq$50 judged PRs after filtering. An interactive version of the leaderboard, including configurable filters and trend visualizations, is available at \url{https://codereview.withmartian.com/}.

\input{tables/top5}

\textbf{Monthly trends.}
F$_1$ rankings are broadly stable across the three-month window. Of the 15~tools, 12 vary by fewer than 5\,pp month-to-month (median absolute change: 1.8\,pp). This period coincided with several major foundation-model releases, and a gradual upward drift may reflect improving base models. However, the drift is modest, suggesting the product harness (prompt design, context selection, comment filtering, etc.) contributes substantially to review quality. The largest shifts are discrete and tool-specific, matching the release events detected in \S\ref{sec:changepoints} (Appendix~\ref{app:changepoints}). If base-model upgrades were the main driver, changepoints would cluster simultaneously across tools; instead they are staggered, indicating that ranking movement more likely comes from product updates. This makes the benchmark directly useful to tool providers as a continuously updated evaluation: teams are tracking their scores over time, verifying that a release improved quality, and diagnosing where (which PR sizes, languages, or severity levels) their tool still underperforms.

\textbf{Isolating the harness contribution.}
The staggered-changepoint evidence above is indirect: it shows that model upgrades alone do not explain ranking shifts. To test this more directly, we built a stripped-down review agent (Minimal Review Agent, or MRA) with a fixed harness of vector-store retrieval, tree-sitter context, linting, and 11~specialized sub-agents, then ran it on the offline benchmark with three different model stacks while holding the harness constant (Table~\ref{tab:mra-inline}).

\begin{wraptable}{r}{0.58\textwidth}
\vspace{-14pt}
\centering
\footnotesize
\setlength{\tabcolsep}{3pt}
\begin{tabular}{llcccc}
\toprule
\textbf{Config} & \textbf{Models (orchestrator + subs)} & \textbf{P} & \textbf{R} & \textbf{F$_1$} & \textbf{Cost} \\
\midrule
mra-openai & GPT-5.5 + GPT-5.4-mini & .166 & .350 & .225 & \$87 \\
mra-mix & GPT-5.5 + Haiku~4.5 & .166 & .350 & .225 & \$491 \\
mra-claude & Sonnet~4.6 + Haiku~4.5 & .157 & .350 & .217 & \$408 \\
\midrule
\multicolumn{2}{l}{\emph{Commercial tools (median)}} & .445 & .467 & .440 & --- \\
\bottomrule
\end{tabular}
\caption{MRA scores on the offline benchmark (50~PRs, Opus 4.5 judge). Same harness, different model families; they achieve identical aggregate recall but find different subsets of issues. Details in Appendix~\ref{app:mra}.}
\label{tab:mra-inline}
\vspace{-12pt}
\end{wraptable}
All three find the same number of golden issues (48) but different subsets (per-PR counts differ on 34 of 50~PRs; SE on recall is $\pm$4.1\,pp). The identical aggregate recall despite per-PR variation suggests MRA's fixed retrieval pipeline determines the scope of what is findable, while the model may determine which issues within that are caught. We frame this as a hypothesis specific to MRA's architecture rather than a universal property of multi-agent systems.

The precision gap to commercial tools (~16\% vs.\ median 44.5\%) confirms our MRA harness lacks the maturity of production systems; this experiment isolates model contribution, not harness quality. Cost varies widely (driven by thinking tokens and tool-call volume) with no quality gain.

For comparison, we ran \textbf{MIRA}\footnote{\url{https://github.com/miracodeai/mira}}, a streamlined single-pass system, on the same offline set with four model families. MIRA fetches the PR diff through the GitHub API, and sends the diff and surrounding context to the model. In this setting, recall spans 15.4\,pp and precision spans 29.8\,pp (Appendix~\ref{app:mra}). The substantial contrast suggests that sensitivity to model choice depends on review architecture. Separately, the Kodus benchmark\footnote{\url{https://www.codereviewbench.com/}} uses an open-source multi-agent system and reports a score spread of only ${\sim}3$\,pp across eight model families, providing a third data point consistent with architecture shaping how strongly model choice appears in benchmark scores.

\textbf{What affects scores.}
PR size is the strongest predictor of quality: F$_1$ drops nearly 28\,pp from small ($\le$50 lines) to extra-large ($\ge$1{,}000 lines) PRs, driven almost entirely by recall collapse (Figure~\ref{fig:size} in Appendix~\ref{app:size-degradation}). This effect is uniform across tools, indicating that size degradation is a property of the task rather than a tool-specific weakness. Other dimensions (severity, language, domain) matter less; tools vary more within a stratum than strata differ from one another (Appendix~\ref{app:stratification}). Review speed does not predict quality either ($\rho = 0.38$, $p = 0.06$; Appendix~\ref{app:speed-quality}), likely because parallel agents make latency a poor proxy for depth and because ingesting more context can hurt given size-degradation. These scores reflect open-source usage only and should not be extrapolated to enterprise settings where stricter review policies and security-sensitive code are absent from the public event stream.

\textbf{Engagement varies by deployment context, not quality.}
Across unfiltered merged PRs, developer engagement after a bot review ranges from  $\sim$32\% to $\sim$61\% across tools (Figure~\ref{fig:usage} in App.~\ref{app:usage-patterns}). Low-engagement tools typically run inside agentic pipelines; high-engagement tools are standalone reviewers inviting explicit responses. This is a deployment-context distinction, not a quality one, and justifies filtering on engagement (\S\ref{sec:quality}): without it, the 28\,pp spread in raw engagement would dominate rankings.

\textbf{The precision--recall frontier.}
AI code review exhibits a real precision--recall frontier (Figure~\ref{fig:pr-tradeoff} in Appendix~\ref{app:pr-tradeoff}). Commenting more aggressively improves coverage but surfaces lower-confidence suggestions that developers ignore. We hypothesize this is partly contextual misalignment: tools that understand which issues a project tends to act on could prioritize better. Tool selection is fundamentally about where on this frontier a team wants to sit. The choice of $\beta$ alone reshuffles the top five among 10 of 15~tools (Table~\ref{tab:top5}), and filter changes reshuffle them further---indicating tools occupy distinct positions on this frontier rather than clustering around a single optimum.

\section{What This Benchmark Unlocks}
\label{sec:applications}
\input{benchmark_unlocks}

\section{Related Work}
\label{sec:related}
\input{related_work}

%% file: tables/top5.tex
% \begin{table}[t]
% \centering
% \small
% \begin{tabular}{clrclrclr}
% \toprule
% & \multicolumn{2}{c}{\textbf{Precision}} &
% & \multicolumn{2}{c}{\textbf{F$_{0.5}$}} &
% & \multicolumn{2}{c}{\textbf{F$_1$}} \\
% \cmidrule{2-3} \cmidrule{5-6} \cmidrule{8-9}
% 1 & Cubic Dev AI       & 75.6 &
%   & Cubic Dev AI       & 67.3 &
%   & Cubic Dev AI       & 57.8 \\
% 2 & Greptile           & 72.4 &
%   & Greptile           & 60.9 &
%   & Gemini             & 52.7 \\
% 3 & Cursor             & 68.7 &
%   & Gemini             & 59.4 &
%   & CodeRabbit         & 52.3 \\
% 4 & Graphite           & 68.3 &
%   & Cursor             & 57.6 &
%   & Claude             & 51.9 \\
% 5 & ChatGPT Codex      & 66.2 &
%   & Claude             & 56.2 &
%   & Sourcery           & 51.6 \\
% \midrule
% & \multicolumn{2}{c}{\textbf{F$_2$}} &
% & \multicolumn{2}{c}{\textbf{Recall}} & & & \\
% \cmidrule{2-3} \cmidrule{5-6}
% 1 & CodeRabbit         & 50.8 &
%   & CodeRabbit         & 49.9 & & & \\
% 2 & Cubic Dev AI       & 50.6 &
%   & Cubic Dev AI       & 46.7 & & & \\
% 3 & Sourcery           & 48.3 &
%   & Sourcery           & 46.3 & & & \\
% 4 & Claude             & 48.2 &
%   & Claude             & 46.0 & & & \\
% 5 & Gemini             & 47.4 &
%   & Gemini             & 44.4 & & & \\
% \bottomrule
% \end{tabular}
% \caption{Top five tools under each scoring metric
%   (Mar~21--Apr~20, 2026, default filters).  Rankings shift
%   across metrics: 9 of 14~tools appear in at least one top
%   five.  All values are percentages.}
% \label{tab:top5}
% \end{table}

\begin{table}[t]
\centering
\small
\begin{tabular}{clrclrclr}
\toprule
& \multicolumn{2}{c}{\textbf{Precision}} &
& \multicolumn{2}{c}{\textbf{F$_{0.5}$}} &
& \multicolumn{2}{c}{\textbf{F$_1$}} \\
\cmidrule{2-3} \cmidrule{5-6} \cmidrule{8-9}
1 & Cubic Dev AI       & 75.5 &
  & Cubic Dev AI       & 67.2 &
  & Cubic Dev AI       & 57.6 \\
2 & Greptile           & 72.6 &
  & Greptile           & 60.9 &
  & Qodo               & 53.0 \\
3 & Cursor             & 68.6 &
  & Gemini             & 59.5 &
  & Gemini             & 52.8 \\
4 & Graphite           & 68.3 &
  & Cursor             & 57.5 &
  & CodeRabbit         & 52.3 \\
5 & ChatGPT Codex      & 66.1 &
  & Qodo               & 57.3 &
  & Claude             & 51.9 \\
\midrule
& \multicolumn{2}{c}{\textbf{F$_2$}} &
& \multicolumn{2}{c}{\textbf{Recall}} & & & \\
\cmidrule{2-3} \cmidrule{5-6}
1 & CodeRabbit         & 50.8 &
  & CodeRabbit         & 49.9 & & & \\
2 & Cubic Dev AI       & 50.5 &
  & Qodo               & 47.2 & & & \\
3 & Qodo               & 49.3 &
  & Cubic Dev AI       & 46.6 & & & \\
4 & Sourcery           & 48.4 &
  & Sourcery           & 46.4 & & & \\
5 & Claude             & 48.2 &
  & Claude             & 46.0 & & & \\
\bottomrule
\end{tabular}
\caption{Top five tools in the online benchmark under each scoring metric
  (Mar~21--Apr~20, 2026, default filters).  Rankings shift
  across metrics: 10 of 15~tools appear in at least one top five. All values are percentages. Qodo scores use the post-update version only (identified by an internal version marker); the default dashboard reports the combined score until the legacy version is fully deprecated.}
\label{tab:top5}
\end{table}

%% file: benchmark_unlocks.tex
Beyond evaluating review tools, the benchmark's design (daily refresh, paired suggestion--action labels, cross-validation against an independent benchmark) enables two further uses:

\paragraph{Training corpus for open-source models:} Open-source review models lack a large corpus pairing comments with whether developers acted on them. \textsc{CodeReviewBench}'s accept/ignore labels can fill this gap if they carry learnable signal, which we test by LoRA fine-tuning Qwen2.5-Coder-7B-Instruct under a within-DPO contrast: \texttt{filtered} uses real labels, \texttt{inverted} swaps \texttt{chosen} and \texttt{rejected} within each pair. We run this on two pipelines: \textbf{PR-level} (full PR diff) and \textbf{hunk-level} (commented hunk plus surrounding lines) at equal training compute. \texttt{filtered} beats \texttt{inverted} by 6--9 pp $F_1$ on both, and PR-level \texttt{filtered} beats the base model by 3.3 pp, above the 1.8 pp median month-to-month drift in \S\ref{sec:findings}. The signal is attributable to the labels themselves. Details in App.~\ref{app:finetuning}; sample at \url{https://huggingface.co/datasets/code-review-bench/code-review-bench}.

\paragraph{Toward automating code review:}
\label{sec:automation} Full automation of code review requires removing the human reviewer from the loop without quality degrading, and the gap varies sharply by the kind of code reviewed. The dominant axis is PR size: F$_1$ drops 28\,pp from small to extra-large PRs (\S\ref{sec:findings}). Severity looks like a second axis --- low-severity issues show lower recall --- but this reflects developer preference rather than tool capability: low-severity comments are dismissed inconsistently. Where a team sits on the automation frontier depends on its codebase and risk tolerance; the configurable leaderboard (\S\ref{sec:findings}) is the instrument for locating that point. Closing the gap entirely also requires using the review signal to train the code generators whose output is being reviewed. For unit-testable code, hidden tests serve as the verifier; for everything else, the verifier has to be a code reviewer. The benchmark does not provide a reward signal directly --- online behavioral signals are asynchronous with code generation. Using it as RL reward would require a learned proxy that predicts how a developer would act at generation time, which we leave to future work.

%% file: related_work.tex
\textbf{Benchmarks for code generation}: 
SWE-bench~\citep{jimenez2024swebench} and its successors (Verified~\citep{openai2024swebenchverified}, Pro~\citep{swebenchpro2025}, Rebench~\citep{badertdinov2025swerebench}) share an objective verifier---hidden unit tests. They have evolved primarily to combat contamination: Verified was retired after training-set leakage~\citep{openai2026whyweretire}, Pro uses copyleft licensing as a barrier, and Rebench continuously refreshes tasks. {\benchmarktext} adopts the continuous-update mechanism and extends it to a setting where no objective verifier exists.
 
\textbf{Benchmarks for AI code review}: 
Prior work relies on curated gold sets: Greptile's 50-PR benchmark~\citep{greptile2025} and Augment's expansion with discussion of incompleteness~\citep{augment2025}. Vendor-run evaluations consistently rank their own tool first~\citep{deepsource2026blog}. No prior benchmark uses revealed developer behavior as the primary signal for AI code review.
 
\textbf{Behavioral and revealed-preference evaluation}: 
Chatbot Arena~\citep{chiang2024chatbotarena} collects pairwise human preferences over model outputs. Our setting differs in two ways: the behavioral signal is passive (developers act naturally, not via explicit voting) and publicly observable (GitHub events vs.\ private IDE sessions).

\section{Limitations}
\label{sec:limitations}
\input{limitations_conclusion}

%% file: limitations_conclusion.tex
\textbf{Imperfect ground truth.}
Revealed preferences are an imperfect proxy for review quality. A suggestion a developer valued but chose not to act on is scored identically to a wrong one; suggestions addressed in follow-up PRs are invisible; issues missed by both developer and reviewer are never counted. Expert-assessed accuracy and automated precision align but are not equivalent (\S\ref{sec:human-validation}). The benchmark trades ground-truth precision for scale and Goodhart resistance; cross-PR causal tracing would narrow this gap.

\textbf{Judge variance.}
The LLM judge adds measurement noise. Cross-judge agreement is high at the ranking level (Fleiss's $\kappa = 0.86$; App.~\ref{app:judge-validation}), an alternative methodology (Comment Action Rate) produces near-identical mean scores, and the judge's matches generally agree with experts (\S\ref{sec:human-validation}), but PR-level scores can still differ. On large diffs, the judge and the tools it evaluates face similar context limits, so both may miss issues without exposing the gap; more broadly, the judge may share systematic blind spots with the tools that cross-judge agreement would not catch. Periodic re-evaluation against fresh expert judgments would help detect drift and bound these effects.

Teams self-select into tools for reasons that correlate with review culture and project maturity, so scores reflect the tool as deployed rather than capability in isolation. DML controlling for nine observable confounders preserves rankings ($\rho = 0.86$; App.~\ref{app:dml}), suggesting no single confounding direction dominates, but unobserved confounders likely remain.

\textbf{Goodhart resistance is not absolute.}
Manipulating scores would require changing how thousands of independent developers respond to suggestions, but resistance is a design goal rather than a guarantee. We identify five plausible vectors: selective deployment in favorable repositories, optimizing for the judge, flooding reviews with low-effort suggestions, creating repositories to suppress competitors' scores, and manipulating known measurement windows. Existing defenses include concentration caps, engagement and team-repository filters, continuous measurement, cross-judge checks, and human validation; Appendix~\ref{app:goodhart} maps each vector to its countermeasures. The list is not exhaustive; we will keep monitoring and add defenses as new strategies surface.

\textbf{Enterprise generalizability.}
The benchmark operates exclusively on public GitHub repositories. Enterprise codebases may differ in review policy, security requirements, languages and frameworks, repository scale, and how developers engage with bot suggestions. Stricter filters requiring multiple contributors and a non-author human reviewer approximate some of these conditions, and rankings remain stable ($\rho \geq 0.78$) on the 9.3\% of PRs (${\sim}$48K) they retain (App.~\ref{app:engagement-sensitivity}). This does not substitute for direct validation on private repositories, e.g., comparing matched public and private deployments using anonymized organization data.

\section{Conclusion}
We presented \benchmark, a paired online-offline benchmark for AI code review where the online benchmark is built on a different reward signal: revealed developer preferences, mined daily from the public GitHub event stream, and cross-validated against a 50-PR offline benchmark scored by expert annotators. The validation establishes that the behavioral signal is reliable, replicable, and robust, and that its precision rankings agree with expert annotations. The paired benchmark helps in addressing the limitations of entirely offline benchmarks, which often go stale. The dataset we provide is the largest open corpus pairing review comments with developer responses, and is already useful to tool providers as a continuously updated evaluation of their tools and for diagnosing where the tools fail. The benchmark also suggests that the performance of code review tools depends strongly on the harness design, as a minimal code review agent built around a simple harness and different frontier LLMs as bases showed no detectable recall difference in our setting. An early fine-tuning experiment shows the labels carry signal a model can learn from. Beyond its current applications, the benchmark could help a wider audience by providing a large corpus for fine-tuning base LLMs to be code reviewers, motivating the construction of synchronous proxy reward signals for RL-trained coding agents, and helping us measure how close we are to automating code review.

%% file: appendix.tex
\section{Broader impacts}
\label{app:impacts}

CodeReviewBench enables transparent, independent evaluation of AI code-review tools, reducing the field's reliance on vendor-published benchmarks where the publishing vendor tends to win. The dataset supports open-source review-model development, lowering the barrier to entry for smaller teams. The continuous-refresh design lets the benchmark track tools as they evolve, rather than going stale at the cadence of yearly curation passes.

Three risks deserve naming. First, rankings are snapshots in a specific deployment context (open-source repositories, particular language and domain mixes); they may not transfer to enterprise settings with different review cultures, and treating them as universal quality judgments would misrepresent what they measure. Second, the methodology itself could become a target: if vendors optimize directly for revealed-preference signal, the benchmark's resistance to Goodharting (which depends on the difficulty of changing real developer behavior at scale) is weakened though not eliminated. Third, behavioral signals encode whatever biases exist in developer behavior, including biases across coding styles, languages, or contributors --- the benchmark surfaces these rather than correcting for them.

We mitigate the first risk by exposing all rankings as configurable rather than aggregate, encouraging teams to evaluate against their own workflow. The second risk is partially addressed by the open methodology: a vendor that games the benchmark becomes visible to anyone running the analysis. The third risk is harder to address structurally and is an important direction for follow-up work.

\section{Online--offline structural contrasts}
\label{app:contrasts}

Table~\ref{tab:contrasts} summarizes the structural differences between the online and offline components of Code Review Bench. Each component's blind spot is the other's strength; disagreements between the two are diagnostic rather than problematic.

\input{tables/online-offline-contrasts}

\noindent The key consequence is that precision is the most portable metric between settings: its rankings show positive agreement (Spearman $\rho = 0.69$, $p = 0.019$). Both settings penalize low-value suggestions, but they do not use identical ground truth. F$_1$ does not transfer ($\rho = -0.05$) because the P/R regime difference inflates different sides of the score. Where the two benchmarks \emph{agree} on a tool's scores, it confirms they track the same underlying capability. Where they \emph{disagree}, the gap distinguishes capability problems (low offline = tool misses real issues) from alignment problems (low online = tool finds real issues but developers don't engage).

\section{F$_\beta$ sensitivity analysis}
\label{app:beta}

\subsection{Within-benchmark ranking sensitivity}

Table~\ref{tab:beta-sensitivity} shows how tool rankings shift as $\beta$ moves from 0.5 (precision-weighted) to 2.0 (recall-weighted). All scores use the engaged+capped filter tier over the window 2026-03-20 to 2026-04-20.

Every tool in the current dataset has measured $P > R$ (\S\ref{sec:metrics}).  Tools with a narrow precision--recall gap (e.g.\ CodeRabbit, $P - R = 0.05$) are relatively insensitive to~$\beta$; tools with a wide gap (e.g.\ Greptile, $P - R = 0.35$) swing sharply. The Spearman rank correlation between F$_1$ and F$_2$ rankings is $\rho = 0.955$, but between F$_{0.5}$ and F$_2$ it drops to $\rho = 0.273$, confirming that~$\beta$ is not merely a cosmetic knob.
\input{tables/beta-sensitivity}

\subsection{Cross-benchmark metric transfer}

We next recompute F$_\beta$ within each benchmark and correlate tool rankings across the 11~tools they share. Agreement is near zero at F$_1$ but increases as recall receives less weight:

\begin{table}[h]
\centering
\small
\begin{tabular}{lr}
\toprule
\textbf{Metric} & \textbf{Cross-benchmark Spearman $\rho$} \\
\midrule
F$_1$ & $-0.05$ \\
F$_{0.5}$ & $+0.55$ \\
F$_{0.3}$ & $+0.66$ \\
Precision & $+0.69$ \\
\bottomrule
\end{tabular}
\caption{Cross-benchmark rank agreement as precision receives more weight. F$_1$ mixes precision with benchmark-specific recall; the precision endpoint isolates the component most comparable across settings.}
\label{tab:cross-benchmark-beta}
\end{table}

\begin{table}[h]
\centering
\small
\begin{tabular}{lrrl}
\toprule
\textbf{Metric} & \textbf{Spearman $\rho$} & \textbf{$p$} & \textbf{Interpretation} \\
\midrule
Precision & $+0.69$ & $0.019$ & Strong agreement \\
P/R Ratio & $+0.60$ & $0.051$ & Trade-off preserved \\
Recall    & $+0.53$ & $0.096$ & Directional, not significant \\
F$_1$     & $-0.05$ & $0.873$ & No correlation \\
\bottomrule
\end{tabular}
\caption{Online vs.\ offline rank correlation ($n = 11$ tools). Online P/R $= 1.53$; offline P/R $= 0.78$.}
\label{tab:cross-benchmark}
\end{table}

The distinction is concrete. A tool that emits many low-value style or speculative suggestions is penalized in both settings: developers tend not to act on them online, and expert curators tend not to include them in the offline gold set. Recall has no equivalent shared denominator. A project-specific concern fixed by the PR author contributes to online coverage even if an offline curator would omit it, while an expert-labeled issue contributes to offline coverage even if the author would defer it.

Robustness checks support the precision agreement: leave-one-out correlations remain positive regardless of which tool is removed, and the bootstrap 95\% CI is $[+0.12,+0.94]$. With 11~tools in both benchmarks, the interval is wide, so the result supports broad precision ordering rather than fine-grained claims about adjacent ranks.

\label{app:coverage}
\input{tables/bot-coverage}

\begin{table}[h]
\centering
\small
\begin{tabular}{lll}
\toprule
\textbf{Concern} & \textbf{Tests} & \textbf{Finding} \\
\midrule
Robustness & 3 judge families, alt.\ methodology, 6 filter configs & Rankings invariant \\
Replicability & Split-halves, random day-partitions, day-block bootstrap & $\sim$1\,pp CIs \\
Fairness & DML with 9 confounders, stratified analysis & Rankings preserved ($\rho{=}0.86$) \\
\bottomrule
\end{tabular}
\caption{Internal validation summary. Full details and agreement matrices in Table~\ref{tab:validation-inventory}.}
\label{tab:validation-summary}
\end{table}

\input{tables/validation-inventory}

\section{Judge validation details}
\label{app:judge-validation}
\input{tables/judge-agreement}

\paragraph{Sampling.}
We draw a stratified sample of 180~PRs from the online database under the default quality filters. The sample includes proportional representation of all 15~leaderboard tools, balanced across precision terciles to ensure coverage of high-, mid-, and low-performing tools.

\paragraph{Model stability.}
Three judge models from different families --- GPT-5~Nano (production), Claude~Sonnet~4.5, Gemini~3~Flash --- independently re-run the full pipeline on the same 180~PRs.
On matching alone (shared extractions from steps~1--2), Fleiss's $\kappa = 0.86$ with pairwise Cohen's $\kappa$ of 0.82--0.90.
On the full pipeline (each model runs extraction and matching independently), per-PR F$_\beta$ correlates at Spearman $\rho = 0.69$--$0.81$ across all model pairs ($p \approx 0$).

\paragraph{Methodology robustness (CAR judge).}
An alternative Comment Action Rate (CAR) judge bypasses the extraction step entirely: for each raw bot comment, it examines the post-review diff and decides whether the diff directly addresses the comment's concern. On 172~PRs with scorable output from both methodologies, the two approaches produce nearly identical mean scores (bias $< 0.01$), agree on the binary outcome (suggestion addressed or not) 82\% of the time (Cohen's $\kappa = 0.63$), and correlate at Spearman $\rho = 0.59$ per PR. Full prompt in Appendix~\ref{app:prompts}.
Disagreements are symmetric and cancel at the tool level, confirming that the extraction-and-matching methodology does not introduce systematic directional bias.

\section{Human validation study details}
\label{app:human-validation}

We conducted two complementary annotation tasks using expert annotators with
software engineering experience. Task~1 tests whether the automated judge
matches suggestions to already-extracted developer actions as a human would.
Task~2 tests whether developer action preserves information about independently
assessed comment quality. The released materials include the task inputs,
annotation instructions, and item-level labels.

\subsection{Task~1: Suggestion--action matching}

\paragraph{Sampling and procedure.}
We sampled 125 bot suggestions from 95~PRs and 11~tools, stratified to balance
the automated judge's matched (62) and unmatched (63) decisions. For each
suggestion, annotators saw the suggestion, the list of developer actions
already extracted for the PR, and an optional diff. They did not see the
judge's decision. Annotators answered \emph{yes}, \emph{no}, or
\emph{uncertain}; for yes labels, they also identified the matching action.

\begin{table}[h]
\centering
\small
\begin{tabular}{lrrr}
\toprule
\textbf{Human judgment} & \textbf{Judge: no match} &
\textbf{Judge: match} & \textbf{Total} \\
\midrule
No        & 49 & 7  & 56 \\
Uncertain & 5  & 9  & 14 \\
Yes       & 9  & 46 & 55 \\
\midrule
Total     & 63 & 62 & 125 \\
\bottomrule
\end{tabular}
\caption{Human and automated matching decisions in Task~1. Excluding the 14
uncertain human judgments, agreement is 95/111 (85.6\%), with Cohen's
$\kappa = 0.712$.}
\label{tab:human-matching}
\end{table}

\paragraph{Disagreement analysis.}
Among the 111 definitive judgments, the judge missed nine matches identified
by annotators and produced seven matches annotators rejected. Errors were more
frequent in shorter sampled diffs: 8/28 for diffs with at most 100 lines,
5/40 for 101--250 lines, and 3/43 for 251--399 lines. No sampled diff exceeded
399 lines. Agreement was 84.6\% for judge confidence $\leq0.4$, 81.0\% for
confidence 0.4--0.7, and 89.5\% above 0.7, so disagreement did not vary
monotonically with confidence.

The 16 disagreements were concentrated among CodeRabbit (4), Greptile (3),
Qodo (3), Copilot (2), Claude (2), Cursor (1), and Sourcery (1), and among
documentation (4), bug (3), refactor (3), other (2), style (2), performance
(1), and security (1) suggestions. These subgroup counts are descriptive:
several tools and categories have small samples, and the study was not powered
to compare their error rates.

\paragraph{Scope.}
This task validates the matching stage conditional on the extracted
suggestions and actions. It does not test whether the preceding extraction
stages recovered every suggestion or developer action.

\subsection{Task~2: Independent comment-quality assessment}

\paragraph{Sampling and procedure.}
We sampled 150 bot comments from 120~PRs and 11~tools, covering eight issue
categories. Annotators saw each comment and the full PR diff, but not the
automated judge output, extracted action summaries, or whether the pipeline
counted the comment as addressed. They rated factual accuracy, relevance,
actionability, severity, and overall quality.

\begin{table}[h]
\centering
\small
\begin{tabular}{llrr}
\toprule
\textbf{Dimension} & \textbf{Rating} & \textbf{Count} & \textbf{Percent} \\
\midrule
Factual accuracy & Yes & 36 & 24.0\% \\
                 & Partially & 80 & 53.3\% \\
                 & No & 34 & 22.7\% \\
\midrule
Relevant to PR   & Yes & 138 & 92.0\% \\
                 & Tangential or no & 12 & 8.0\% \\
\midrule
Actionable       & Yes & 148 & 98.7\% \\
                 & Somewhat or no & 2 & 1.3\% \\
\midrule
Overall quality  & Useful ($\geq3$ of 5) & 107 & 71.3\% \\
                 & Low value ($<3$ of 5) & 43 & 28.7\% \\
\bottomrule
\end{tabular}
\caption{Descriptive statistics for the 150 Task~2 annotations.}
\label{tab:human-quality}
\end{table}

\paragraph{Review-time accuracy adjustment.}
The full diff includes commits made after the bot review. Consequently, an
annotator could mark a comment partially accurate because a later commit had
already fixed the issue, even though the issue existed when the bot commented.
We counted a partial label as identifying a real issue only when the
annotator's explanation explicitly attributed the partial rating to a later
commit or hunk in the same PR. The operational rule matched phrases such as
\texttt{later commit}, \texttt{later hunk}, \texttt{already fixed},
\texttt{already addressed}, and \texttt{landed} or \texttt{added later}.
This adjustment applies to 55 of the 80 partial labels. Together with the 36
unqualified yes labels, 91/150 comments (60.7\%) identify real issues at the
time of review.

\begin{table}[h]
\centering
\small
\begin{tabular}{lrrr}
\toprule
\textbf{Tool} & \textbf{$n$} & \textbf{Human accuracy} &
\textbf{Online precision} \\
\midrule
Augment    & 8  & 75.0\% & 57.1\% \\
Claude     & 18 & 61.1\% & 52.4\% \\
CodeRabbit & 32 & 53.1\% & 45.3\% \\
Copilot    & 19 & 63.2\% & 58.5\% \\
Cubic      & 6  & 66.7\% & 63.5\% \\
Cursor     & 7  & 85.7\% & 71.1\% \\
Devin      & 7  & 57.1\% & 56.9\% \\
Gemini     & 15 & 66.7\% & 51.5\% \\
Greptile   & 14 & 71.4\% & 56.8\% \\
Qodo       & 14 & 35.7\% & 48.8\% \\
Sourcery   & 10 & 60.0\% & 55.7\% \\
\bottomrule
\end{tabular}
\caption{Task~2 review-time accuracy and automated online precision by tool.
The per-tool sample sizes support a rank-level validation, not precise
tool-specific human quality estimates.}
\label{tab:human-tool-accuracy}
\end{table}

\paragraph{Agreement with automated precision.}
Across the 11~tools, review-time human accuracy and automated online precision
correlate at Spearman $\rho = 0.688$ (two-sided $p = 0.019$). This indicates
that tools whose comments experts more often judge accurate also tend to score
higher on developer-action precision, although the two signals are not
equivalent.

\paragraph{Scope.}
Task~2 evaluates the quality of comments the bots produced. It does not ask
annotators to independently review each PR, so it cannot estimate human
reviewer recall or serve as a complete human precision--recall baseline.

%% --- Quality Signal ---

\label{app:engagement-raw}
\begin{figure}[h]
\centering
\includegraphics[width=\textwidth]{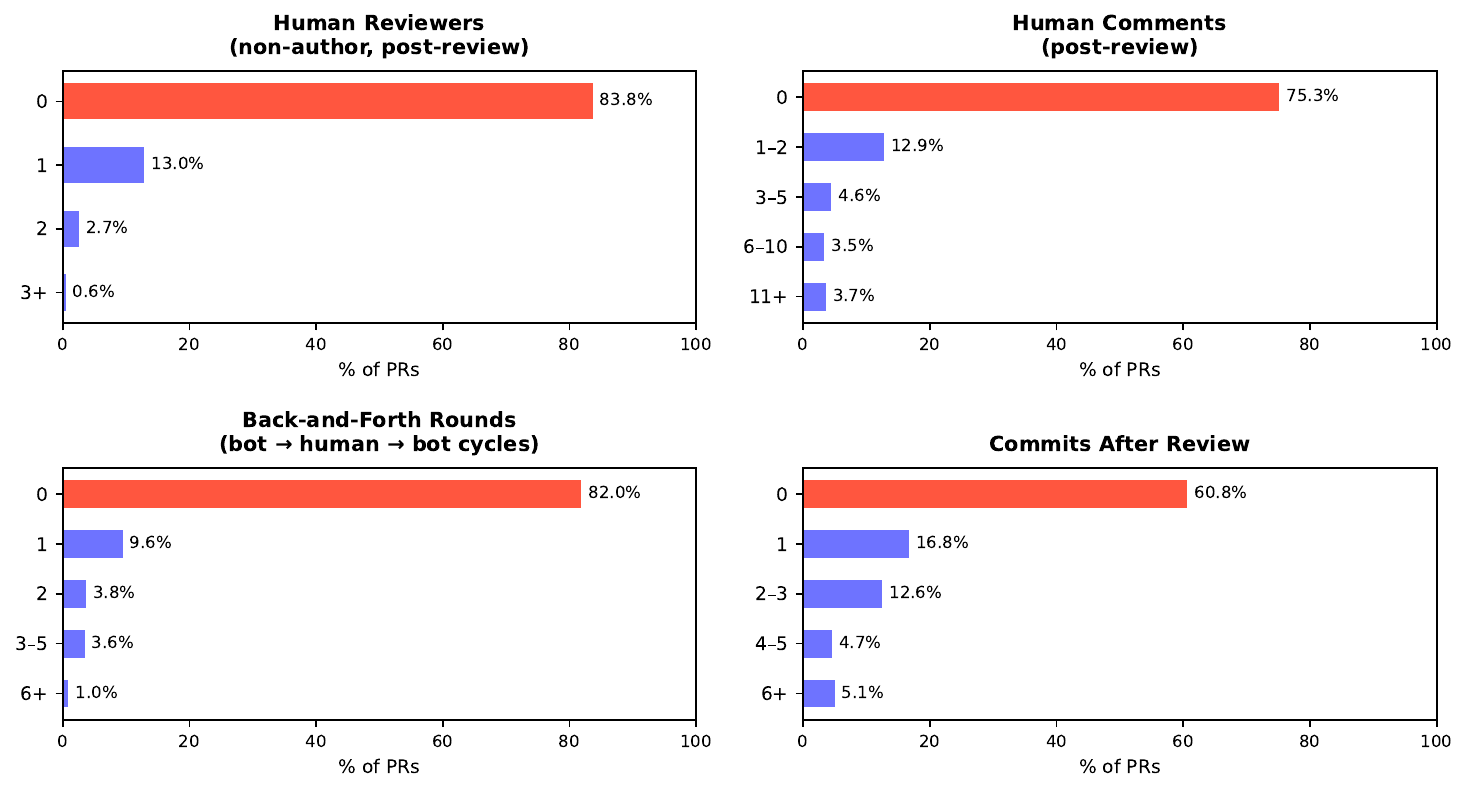}
\caption{Distribution of four engagement signals across 518K merged, bot-reviewed PRs. \textbf{Red bars} show PRs with zero activity for that signal; \textbf{blue bars} show non-zero counts. \emph{Human Reviewers} counts distinct non-author reviewers who commented after the bot's review; 83.8\% of PRs have none. \emph{Human Comments} counts post-review human comments; 75.3\% have none. \emph{Back-and-Forth Rounds} counts bot$\to$human$\to$bot review cycles; 82.0\% have none. \emph{Commits After Review} counts commits pushed after the bot's review; 60.8\% have none. The dominance of zero-activity PRs motivates the quality filtering described in \S\ref{sec:quality}.}
\label{fig:engagement-raw}
\end{figure}

\label{app:filter-convergence}
\begin{figure}[h]
\centering
\includegraphics[width=0.75\textwidth]{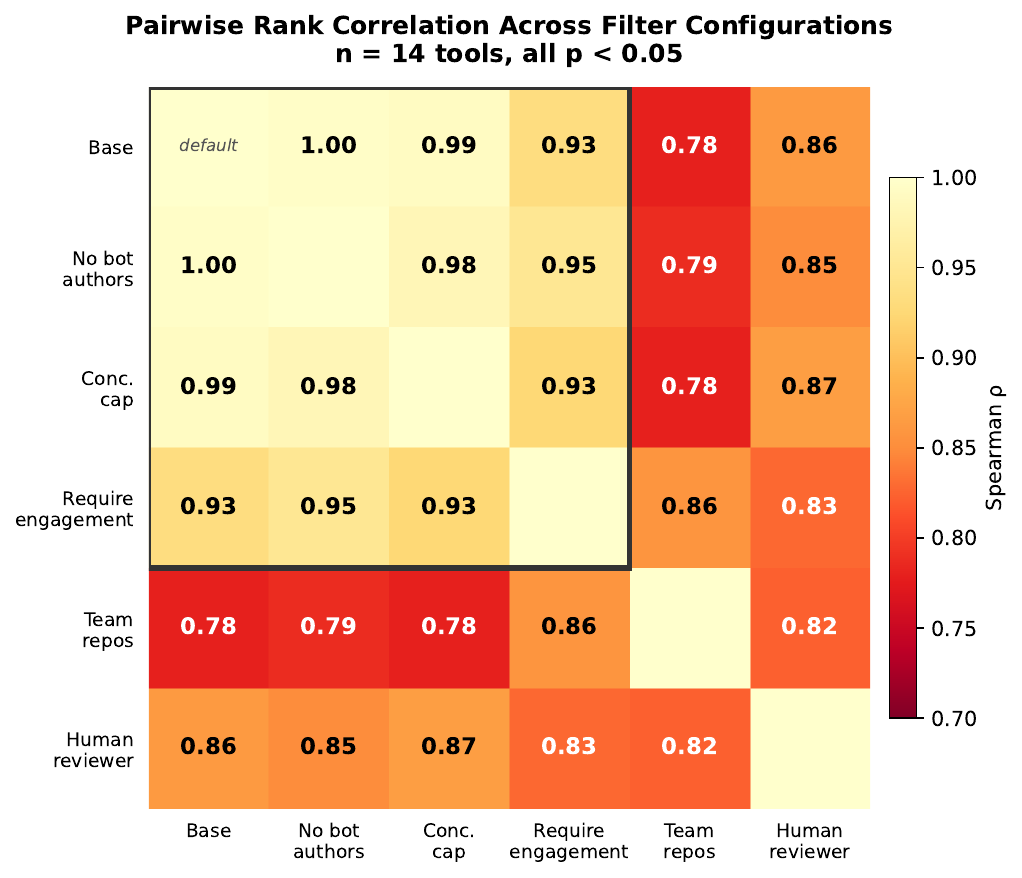}
\caption{Pairwise Spearman $\rho$ of tool F$_1$ rankings across six cumulative filter configurations ($n = 14$ tools, all $p < 0.05$). Within the default-filter steps (boxed), $\rho \geq 0.93$; even base vs.\ strictest, $\rho \geq 0.78$.}
\label{fig:filter-heatmap}
\end{figure}

\label{app:filter-attrition}
\begin{figure}[h]
\centering
\includegraphics[width=\textwidth]{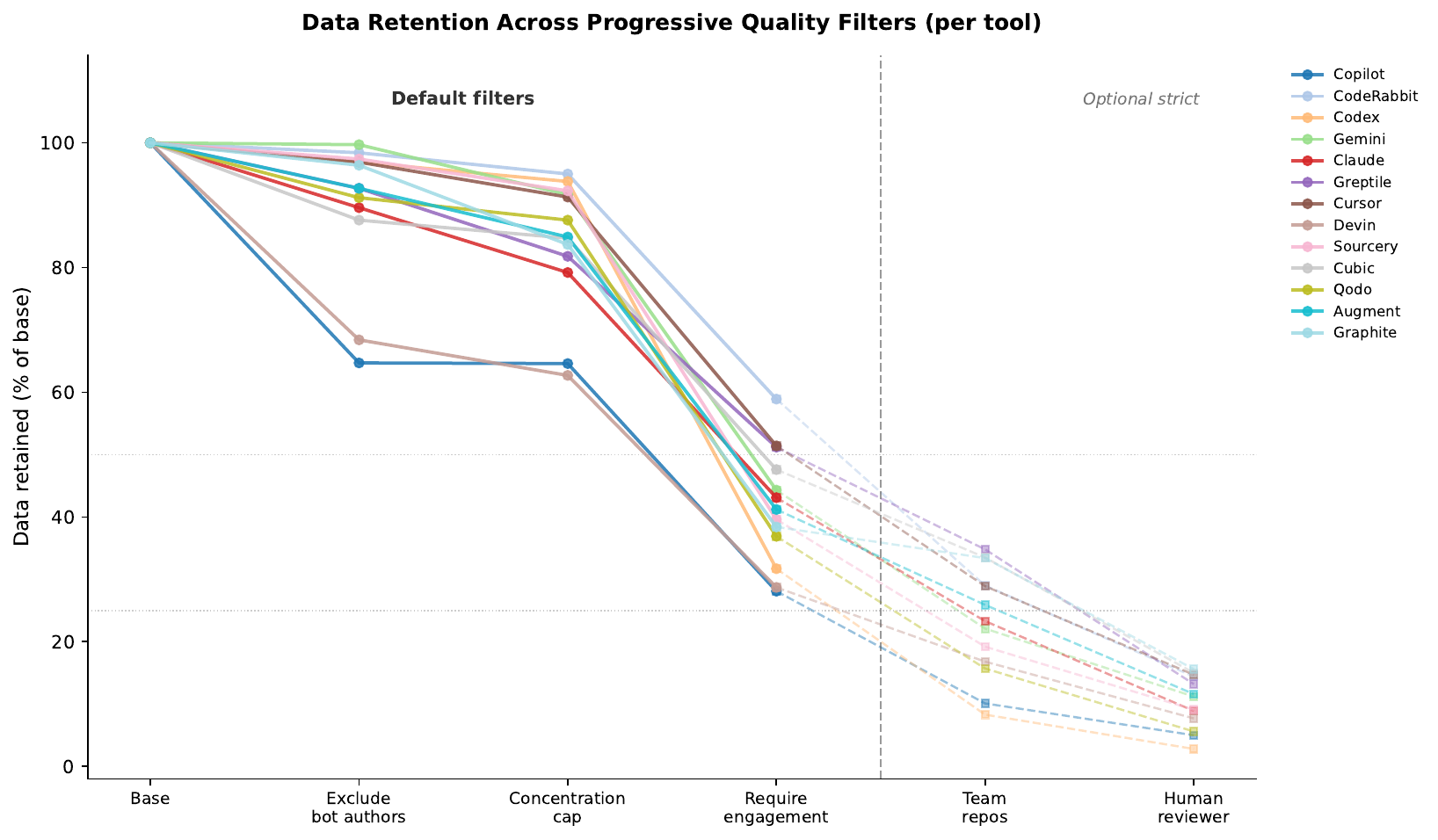}
\caption{Data retention as quality filters are applied cumulatively, per tool. Solid lines show the default filter cascade; dashed lines show the optional strict layers. The engagement filter causes the largest single cumulative drop ($-$42\,pp), reflecting the prevalence of zero-interaction PRs; the team-repository filter removes a further $\sim$21\,pp, consistent with the dominance of solo-developer repositories (see Table~\ref{tab:filters} for exact cumulative retention rates).}
\label{fig:attrition}
\end{figure}

\label{app:usage-patterns}
\begin{figure}[h]
\centering
\includegraphics[width=0.85\textwidth]{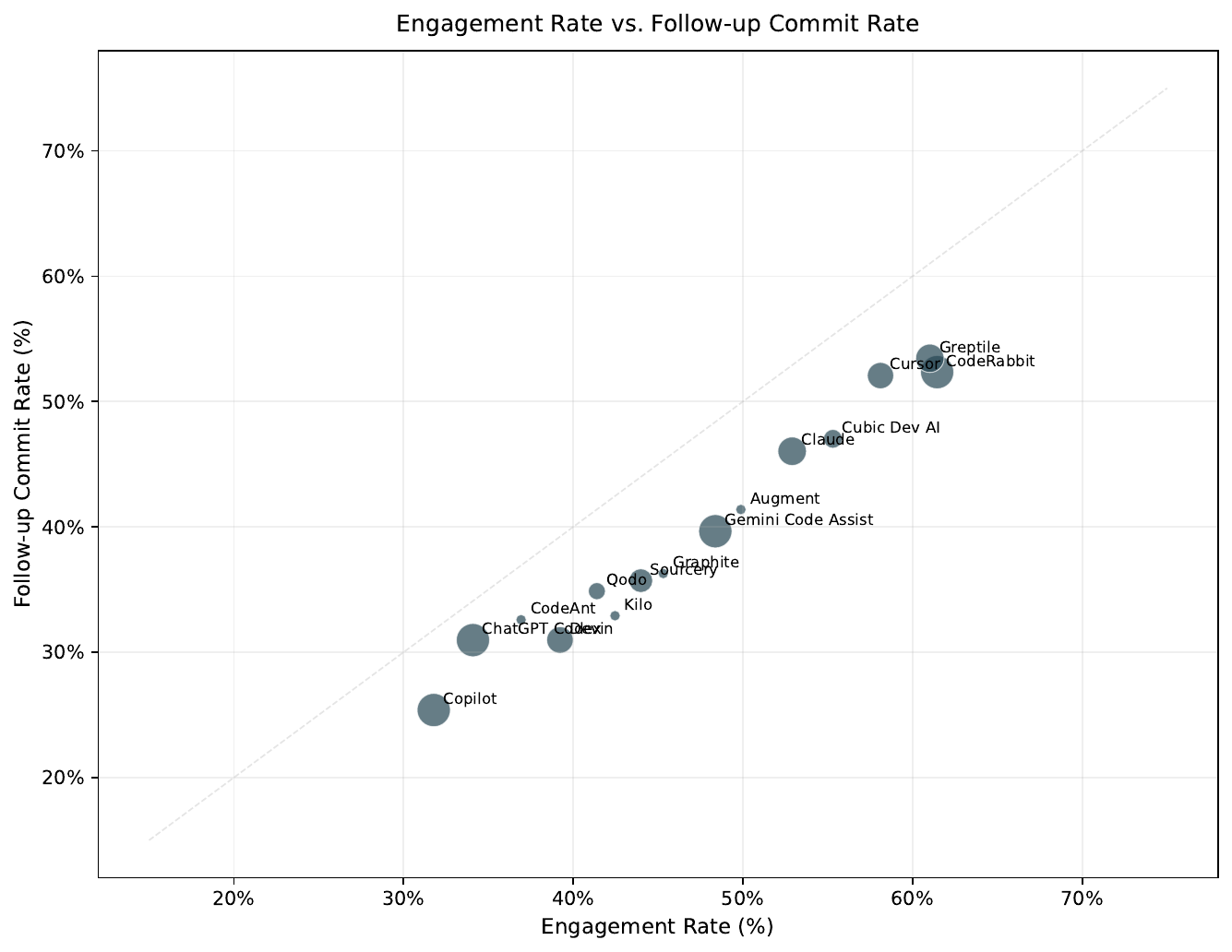}
\caption{Per-tool engagement rate vs.\ follow-up commit rate across all merged PRs (unfiltered, Feb~16--Apr~20, 2026). Marker size scales with PR volume.}
\label{fig:usage}
\end{figure}

\label{app:engagement-corr}
\begin{figure}[h]
\centering
\includegraphics[width=0.75\textwidth]{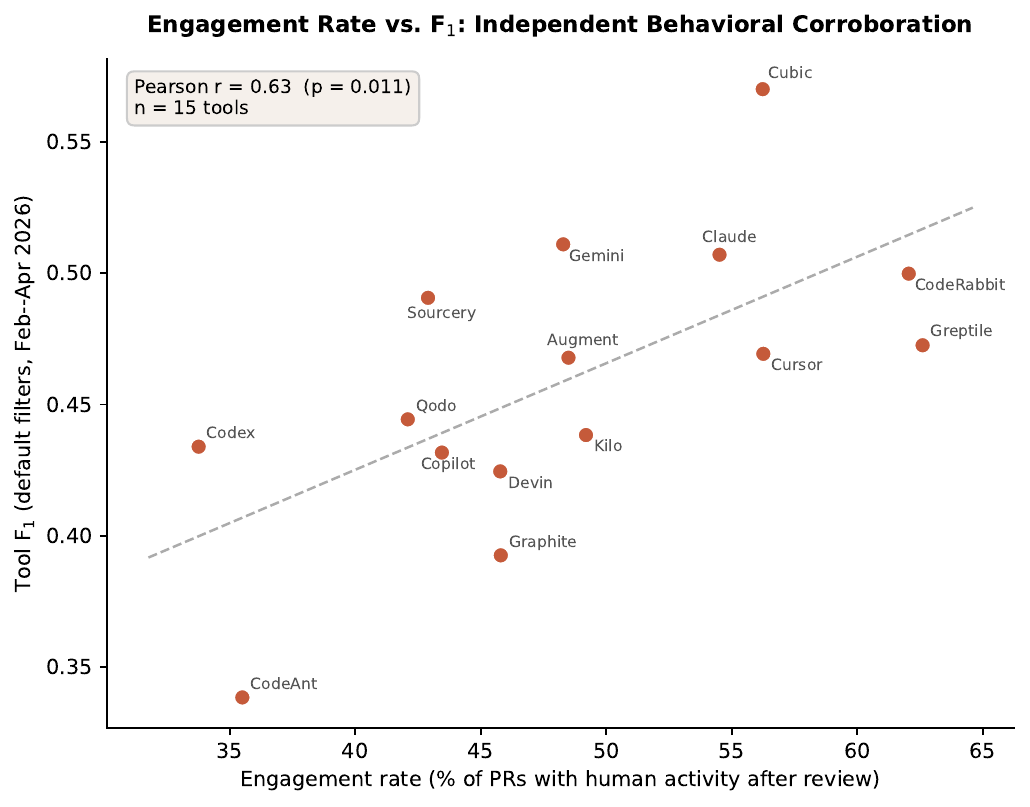}
\caption{Per-tool engagement rate (fraction of PRs with human activity after the bot's review, computed on all filtered PRs before the engagement filter) vs.\ tool F$_1$ (computed on the engagement-filtered set).  Pearson $r = 0.63$, $p = 0.011$, $n = 15$ tools. The two measurements are independent: engagement rate is a behavioral signal from the GitHub event stream; F$_1$ is the LLM judge's assessment of suggestion quality.}
\label{fig:engagement-f1}
\end{figure}

\label{app:temporal-stability}
\begin{figure}[h]
\centering
\includegraphics[width=\textwidth]{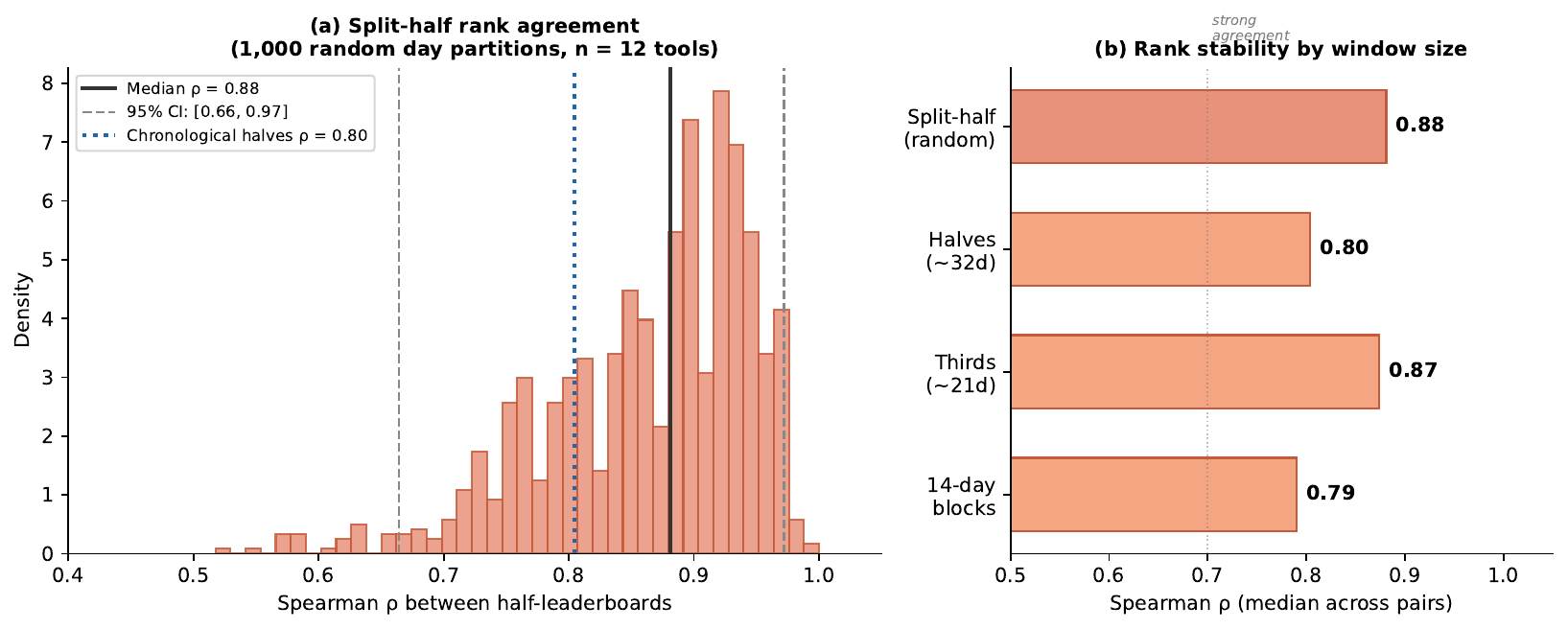}
\caption{Temporal stability of tool rankings (12 tools with $\geq$50 scored PRs, Feb 16--Apr 20, 2026, 64-day window).
  \textbf{(a)}~Spearman $\rho$ from 1{,}000 random day partitions; median $\rho = 0.88$, 95\% CI $[0.66,\;0.97]$.
  \textbf{(b)}~Median $\rho$ across chronological splits at three granularities; all exceed 0.70. The chronological split ($\rho = 0.80$, $p = 0.002$) is a lower bound as it absorbs real product updates mid-window.}
\label{fig:temporal-stability}
\end{figure}

\section{Engagement sensitivity analysis}
\label{app:engagement-sensitivity}

A natural concern is whether the quality filters themselves introduce bias into precision and recall estimates. We test this by varying two key filtering dimensions --- minimum repository contributors and minimum human reviewers --- and observing how aggregate precision and recall shift. All runs use the default window (Feb 16--Apr 20, 2026) with bot-author exclusion, concentration cap, and engagement requirement held constant.
\input{tables/engagement-sensitivity}

\paragraph{Findings.}
Precision is remarkably stable across contributor thresholds (0.586--0.605), exhibiting a shallow U-shape: it dips slightly for mid-sized teams and recovers for large teams ($\ge$6 contributors). Recall declines modestly as the contributor threshold increases (0.415 $\to$ 0.400), consistent with the hypothesis that larger, more experienced teams are more selective about which suggestions they act on --- they may already know the issue or have higher thresholds for what warrants a code change.

The human-reviewer dimension tells a different story. Requiring $\ge$2 human reviewers \emph{lowers} both precision (0.596 $\to$ 0.551) and recall (0.415 $\to$ 0.373). We hypothesize two contributing factors: (1)~PRs that attract multiple reviewers tend to be more complex (larger diffs, cross-cutting concerns), which degrades tool performance as shown in the PR-size analysis (Appendix~\ref{app:size-degradation}); (2)~when multiple humans provide feedback, the developer may address a human reviewer's concern rather than the bot's overlapping suggestion, causing the bot's comment to be scored as unmatched.

The engagement filter itself barely moves precision ($+$0.004) and has no effect on recall, confirming the filter-convergence finding in the main text (\S\ref{sec:quality}): the engagement requirement selects for PRs where signal exists without substantially biasing what signal is measured.

\paragraph{Implications.}
The non-monotonic and dimension-dependent relationships confirm that no single confounding direction dominates. Solo-developer repositories neither systematically inflate scores (by blindly accepting) nor deflate them (by ignoring); similarly, multi-reviewer PRs do not uniformly sharpen signal. Under the strictest configuration (multi-contributor repositories with active human reviewers), tools designed for collaborative workflows gain ground --- Augment, for example, rises to the top, suggesting its suggestions are well-calibrated for teams that actively engage with review feedback. The binary filter approach used in the main paper is a reasonable first approximation, but graduated weighting by signal depth --- assigning more weight to PRs from larger teams with active reviewers --- remains a natural extension for future work.

\section{Causal debiasing via Double Machine Learning}
\label{app:dml}

PRs are not randomly assigned to tools --- teams self-select, and that selection may correlate with repo characteristics, review culture, and project maturity. To estimate whether this selection bias drives our rankings, we apply Double/Debiased Machine Learning~\citep{chernozhukov2018dml} to separate each tool's causal effect on developer action rate from observable confounders.

\paragraph{Method.}
For each tool $d$, we fit a partially linear model $Y = \theta D + g(X) + U$, where $Y$ is per-PR precision, $D$ is a binary indicator for tool $d$ vs.\ all others, and $X$ is a vector of confounders: log diff size, number of human reviewers, human comment count, commits after review, back-and-forth rounds, language, domain, repo activity level, and suggestion count. Using 5-fold cross-fitting, we train gradient-boosted models on held-out folds to predict $Y$ from $X$ (outcome nuisance $\hat{g}$) and $D$ from $X$ (propensity $\hat{m}$), then regress residuals $\tilde{Y} = Y - \hat{g}(X)$ on $\tilde{D} = D - \hat{m}(X)$. The Neyman-orthogonal structure ensures small errors in $\hat{g}$ and $\hat{m}$ bias $\hat{\theta}$ only at second order, yielding $\sqrt{n}$-consistent estimates with valid confidence intervals.

\paragraph{Reading the table.}
The \emph{raw diff} is the naive comparison: mean precision on PRs reviewed by tool~$d$ minus mean precision on PRs reviewed by all other tools. $\hat{\theta}$ is the DML estimate of the causal effect --- how much using tool~$d$ changes precision after stripping out the contributions of diff size, reviewer count, language, domain, and the other confounders listed above. The gap between the two reveals how much of a tool's apparent advantage (or disadvantage) is explained by its deployment context rather than its own quality. SE is the standard error of $\hat{\theta}$; entries marked $^*$ are significant at $p < 0.05$.

\paragraph{Results.}
Table~\ref{tab:dml} reports debiased effects for the 12 tools with $\geq$200 scored PRs. The Spearman rank correlation between raw and debiased effects is $\rho = 0.86$ ($p < 0.001$), confirming that selection bias does not drive rankings. Confounders attenuate raw differences --- the mean absolute shift is 4.4\,pp --- but do not reorder tools.
\input{tables/dml}

Several patterns are noteworthy. For most tools, the debiased effect is smaller in magnitude than the raw difference, meaning part of their apparent advantage or disadvantage is explained by deployment context --- the repositories, team sizes, and review cultures in which they happen to be used. In a few cases, more than half the raw gap is attributable to confounders rather than tool quality. One tool (Copilot) exhibits a sign flip: its raw effect is slightly negative ($-$0.015) because it is disproportionately deployed in solo-developer and low-engagement repositories, but after controlling for this, its causal effect is modestly positive ($+$0.012). The overall ranking, however, is preserved ($\rho = 0.86$), and the top and bottom tools remain clearly separated even after debiasing.

\section{Noise floor}
\label{app:noise-floor}

A benchmark that cannot statistically distinguish between tools is not informative regardless of its other properties. We estimate 95\% confidence intervals on each tool's F$_1$ via day-block bootstrap: resampling whole days with replacement 2{,}000 times and recomputing the weighted-average F$_1$ per tool in each resample.

\begin{figure}[h]
\centering
\includegraphics[width=0.75\textwidth]{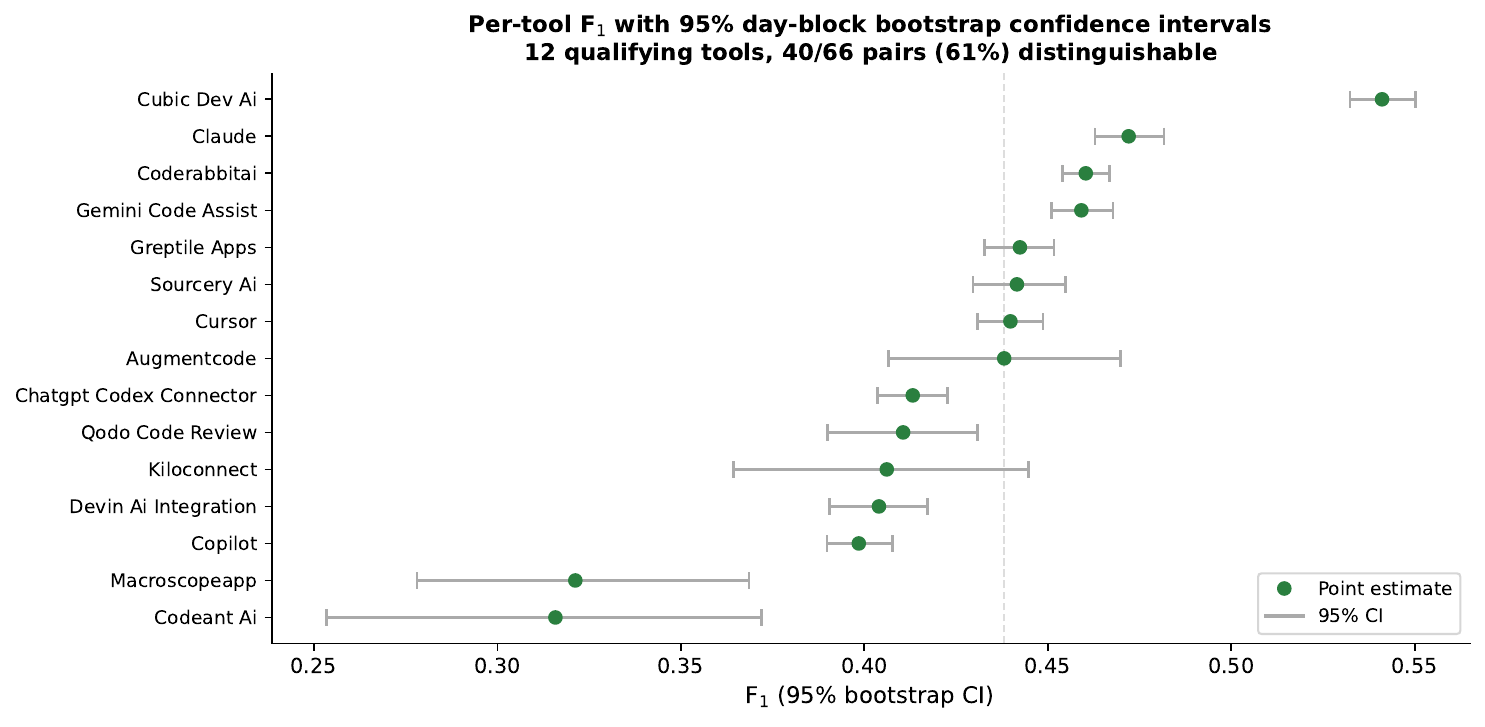}
\caption{Per-tool F$_1$ with 95\% bootstrap confidence intervals
  (2{,}000 day-block resamples, default filters, 12 tools with
  $\geq$30 qualifying days).  Median SE $= 0.005$.  61\% of
  tool pairs (40/66) have non-overlapping CIs.}
\label{fig:noise-floor}
\end{figure}

The median standard error is 0.005 ($\approx$\,1\,pp), meaning a typical tool's F$_1$ is known to within roughly $\pm$1\,pp (Figure~\ref{fig:noise-floor}).  Of the 66 pairwise comparisons among the 12 qualifying tools, 40 (61\%) have non-overlapping 95\% CIs and are therefore statistically distinguishable. The remaining 39\% are tools whose F$_1$ is genuinely close---no benchmark can separate them without substantially more data. Tools with wider CIs (Augment, Qodo, Devin) have fewer qualifying days after engagement filtering, which the CI widths correctly reflect.

%% --- Leaderboard Findings ---

\label{app:pr-tradeoff}
\begin{figure}[h]
\centering
\includegraphics[width=0.85\textwidth]{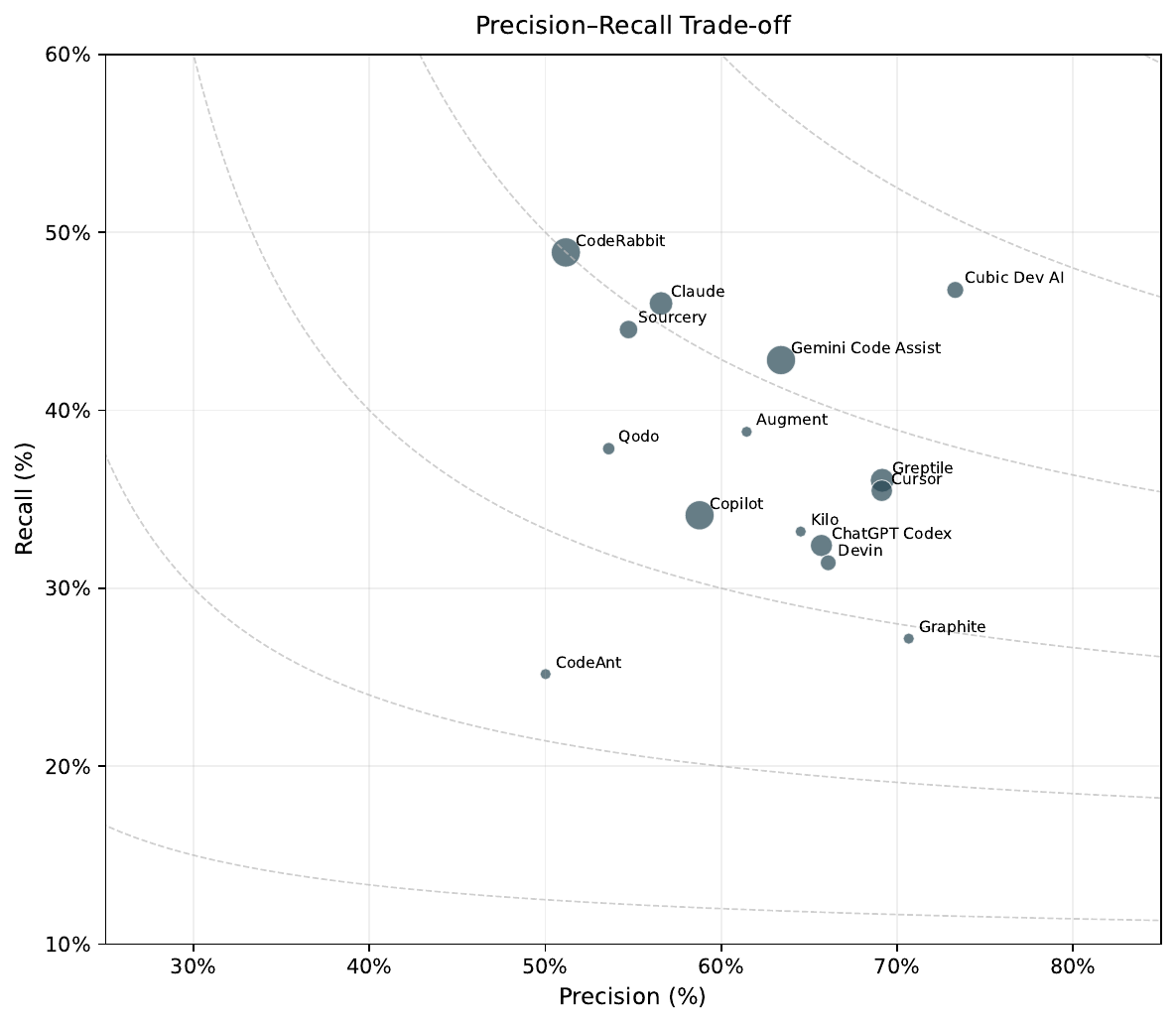}
\caption{Precision vs.\ recall for 15~tools (default filters, Feb 16--Apr 20, 2026). Dashed curves are iso-F$_1$ contours; marker size scales with scored-PR count. No tool occupies the upper-right corner. Median precision across tools is 63.4\%; median recall is 35.5\%. CodeRabbit ranks \#1 in recall (48.9\%) but \#14 in precision (51.2\%); Graphite inverts (\#2 precision, \#14 recall), illustrating the trade-off.}
\label{fig:pr-tradeoff}
\end{figure}

\label{app:size-degradation}
\begin{figure}[h]
\centering
\includegraphics[width=0.85\textwidth]{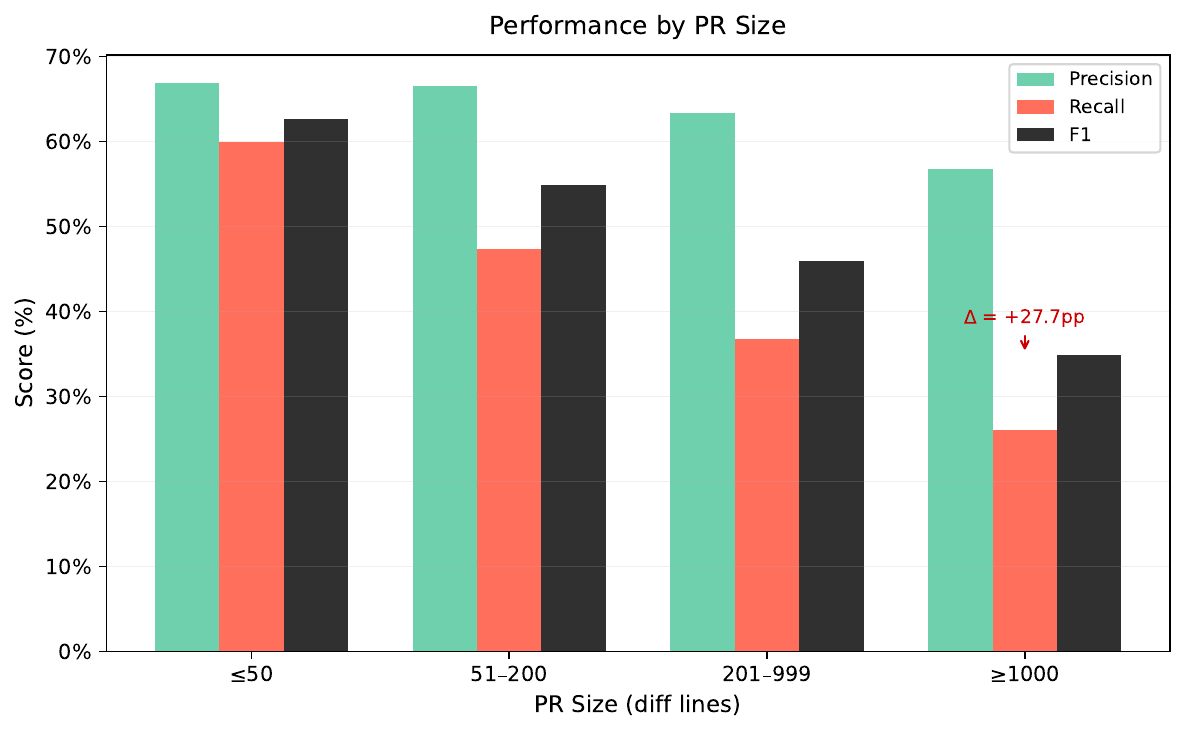}
\caption{Aggregate precision, recall, and F$_1$ by PR size (default filters, 15~tools). F$_1$ drops 27.7\,pp from small ($\le 50$ lines; F$_1$ = 62.5\%) to extra-large ($\ge 1{,}000$ lines; F$_1$ = 34.8\%), driven almost entirely by recall collapse (59.9\% $\to$ 26.0\%). Precision degrades only modestly (66.9\% $\to$ 56.7\%). The effect is uniform across tools.}
\label{fig:size}
\end{figure}

\section{Stratification details}
\label{app:stratification}

\subsection{Severity}

\begin{figure}[h]
\centering
\includegraphics[width=0.85\textwidth]{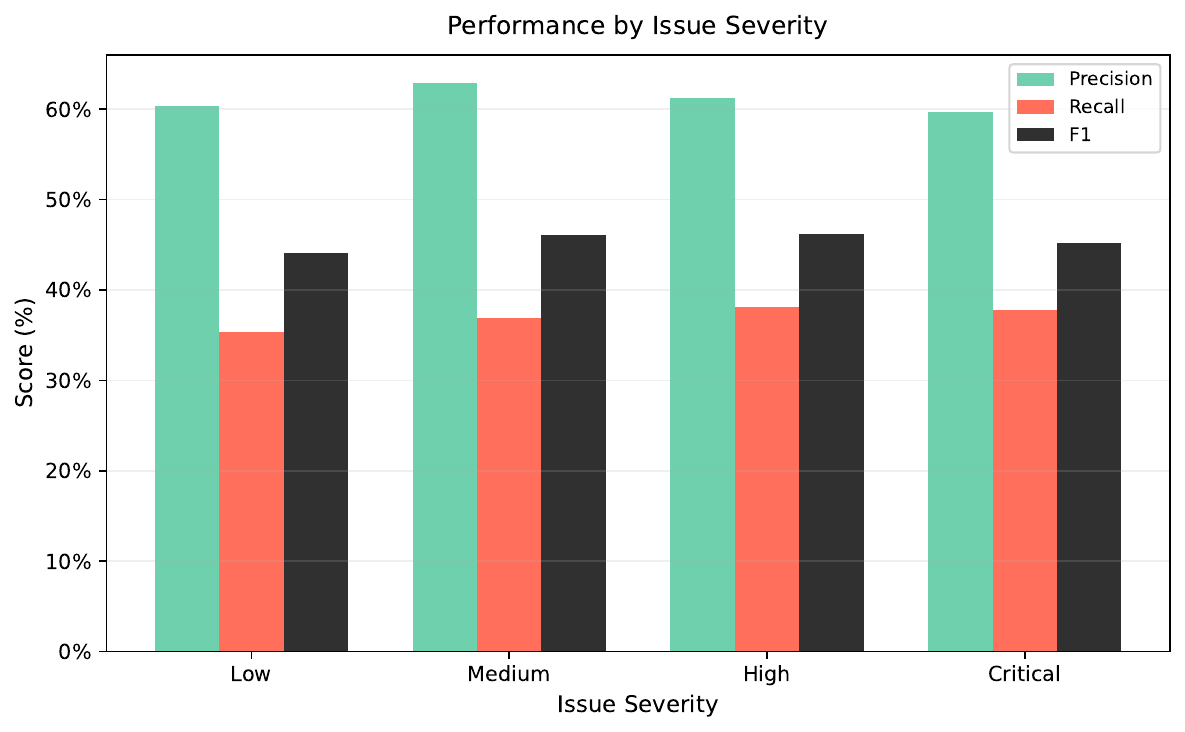}
\caption{Aggregate precision, recall, and F$_1$ by
  LLM-assigned issue severity.  Variation across severity
  levels is small ($<$3\,pp in recall); low-severity
  recall is depressed by developer-preference variance
  rather than tool failure.}
\label{fig:severity}
\end{figure}

Aggregate recall is nearly flat across severity levels: 35.4\% on low-severity issues, 36.9\% on medium, 38.1\% on high, and 37.7\% on critical (Figure~\ref{fig:severity}). The modest 2.7\,pp gap between low and high severity suggests that tools are not meaningfully better at catching critical bugs than minor ones.  Low-severity comments --- style suggestions, naming nitpicks, minor refactors --- are more subject to individual developer preference: some teams welcome them, others dismiss them regardless of correctness. This preference-driven variance in the ground truth dampens measured recall for low-severity issues without implying the tools failed to surface them. Precision is similarly stable (60--63\%).

\subsection{Language and domain}

\begin{figure}[h]
\centering
\includegraphics[width=0.85\textwidth]{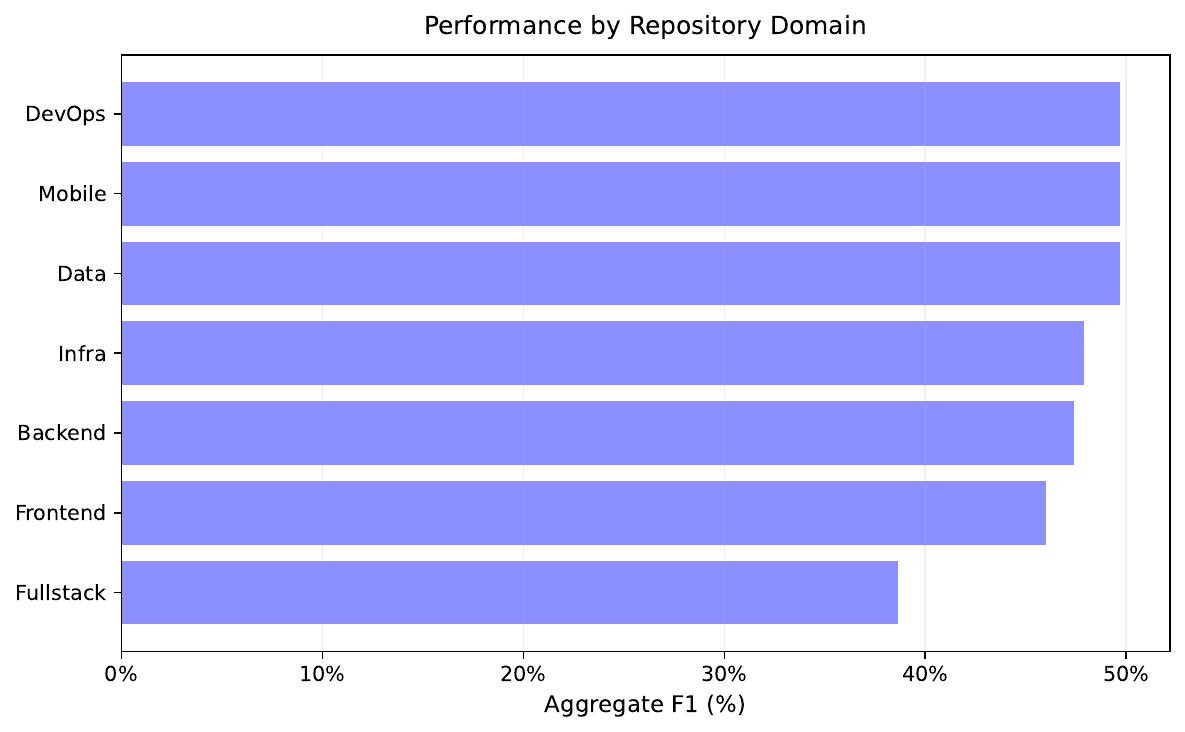}
\caption{Aggregate F$_1$ by repository domain.  Fullstack PRs
  are the hardest; other domains cluster within 1.5\,pp.}
\label{fig:domain}
\end{figure}

Per-domain variation is small: backend, infrastructure, and frontend cluster within 1.5\,pp, while fullstack PRs are the hardest (F$_1 = 38.3\%$), likely because they span more files and mix concerns across the stack (Figure~\ref{fig:domain}).

\begin{figure}[h]
\centering
\includegraphics[width=\textwidth]{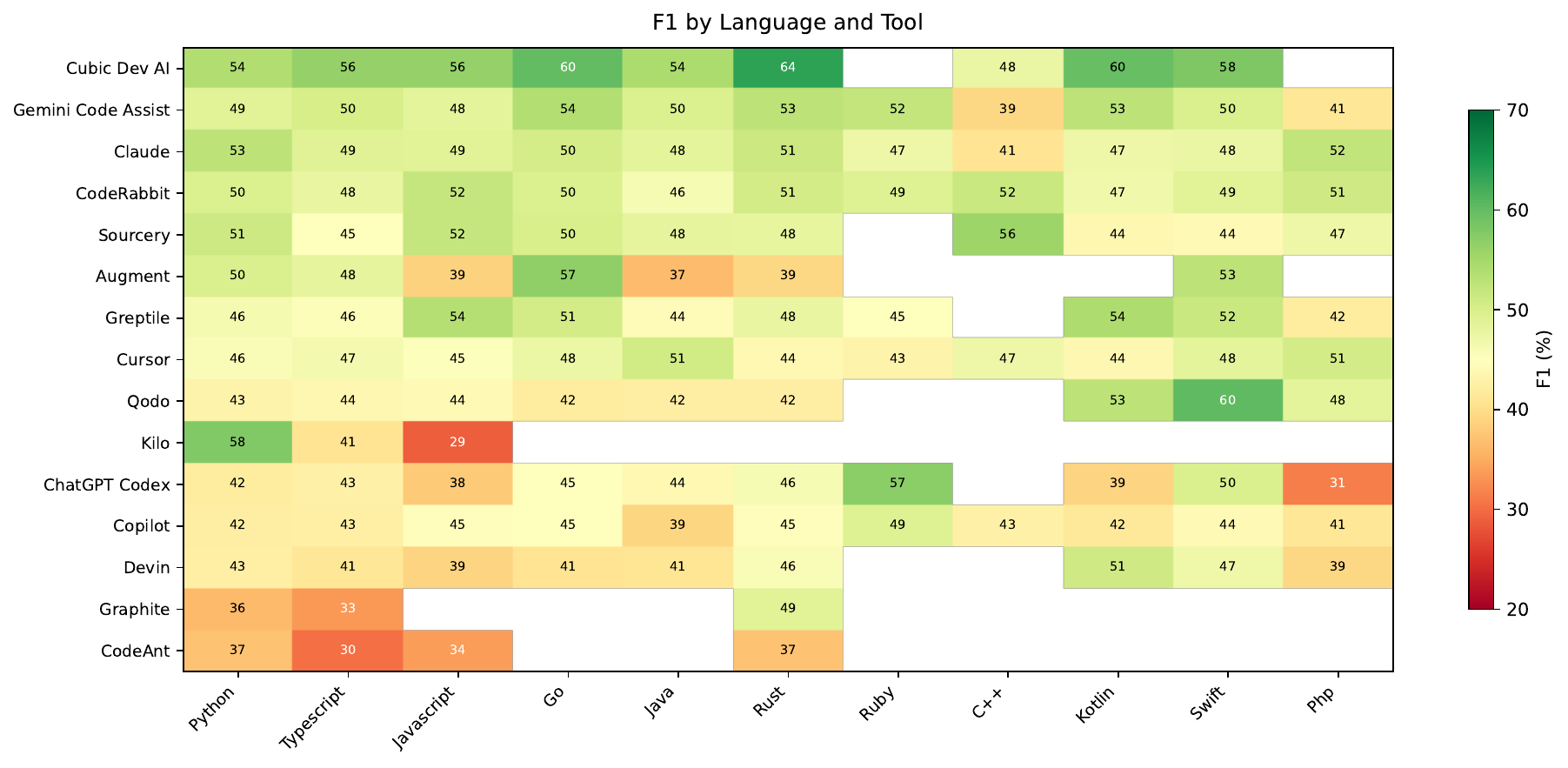}
\caption{F$_1$ by language and tool.  Cells with fewer than
  10~scored PRs are blank.  Between-tool variance within each
  language exceeds between-language variance.}
\label{fig:lang-heatmap}
\end{figure}

Per-language variation (Figure~\ref{fig:lang-heatmap}) is similarly compressed. Go and Ruby show slightly higher mean F$_1$ (${\sim}49\%$) than Python or TypeScript (${\sim}43$--$45\%$), but between-tool variance within each language exceeds between-language variance --- tool choice matters more than language choice.

\subsection{Review volume}

\begin{figure}[h]
\centering
\includegraphics[width=0.85\textwidth]{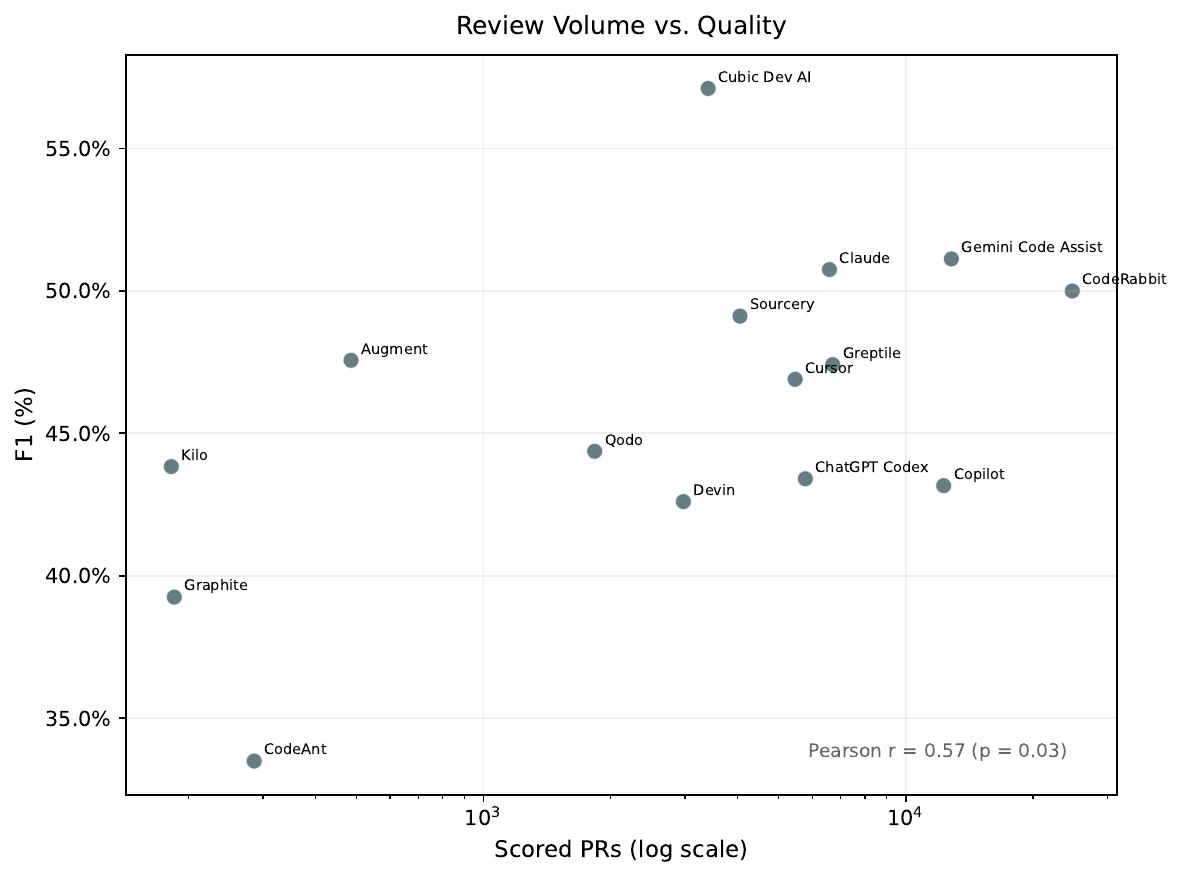}
\caption{Scored-PR volume vs.\ F$_1$ (log-$x$ scale).
  Pearson $r = 0.37$ ($p = 0.17$); not significant.
  Volumes reflect open-source PRs only.}
\label{fig:volume}
\end{figure}

Pearson $r = 0.37$ ($p = 0.17$) between total scored PRs and F$_1$ (Figure~\ref{fig:volume}). The relationship is not significant: several smaller tools achieve top-tier scores on individual metrics despite lower overall volume. The benchmark does not implicitly reward coverage. These volumes reflect open-source usage only; closed-source adoption patterns may differ substantially.

%% --- Offline ---

\label{app:cross-benchmark}
\begin{figure}[h]
\centering
\includegraphics[width=\textwidth]{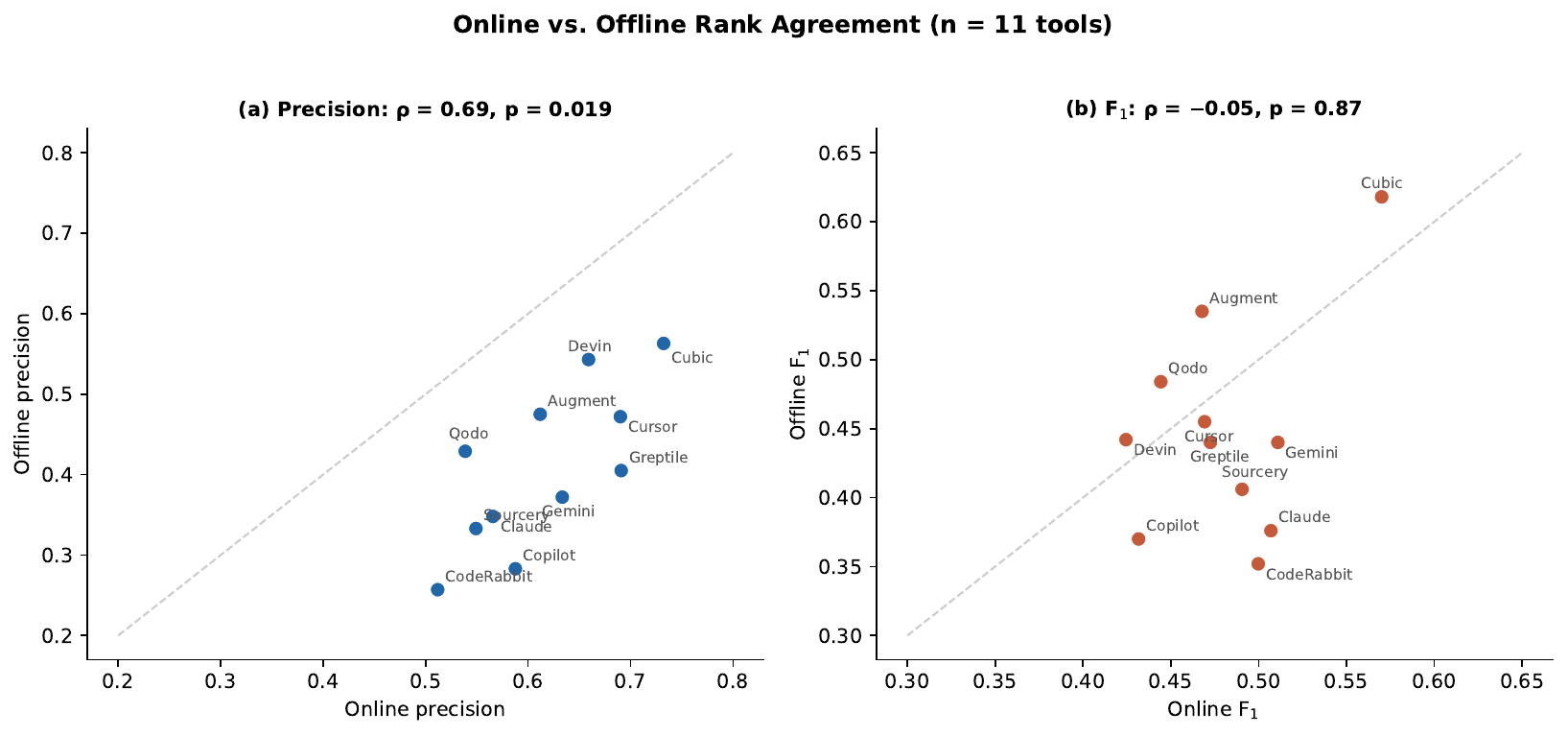}
\caption{Online vs.\ offline rank agreement for 11 tools
  (online: default filters, Feb 16--Apr 21 2026;
  offline: most recent 50-PR run per tool).
  \textbf{(a)}~Precision rankings agree significantly ($\rho = 0.69$, $p = 0.019$); points cluster near the diagonal.
  \textbf{(b)}~F$_1$ rankings show no correlation ($\rho = -0.05$) because the two benchmarks operate in different precision--recall regimes (online P/R $= 1.53$; offline $= 0.78$).}
\label{fig:cross-benchmark}
\end{figure}

\section{Gold-set incompleteness analysis}
\label{app:gold-set-analysis}

The offline gold set extends the 50-PR benchmark originally created by \citet{greptile2025} and subsequently refined by \citet{augment2025}. Our contributions include removing one incorrectly labeled issue, adding severity annotations, formalizing the deduplication pipeline, and evaluating 11 additional tools (along with re-running updated results for the existing tools). The analysis below identifies further incompleteness; we plan to incorporate these findings into the next gold-set revision.

For each of the 50~PRs, the offline benchmark provides two scoring signals: \emph{curated labels} (the current gold set, 136~issues) and \emph{tool output} (the comments each tracked tool produced).

\paragraph{Issues the gold set misses.}
We cluster all false-positive comments per PR across the 20~tools on the public leaderboard by TF-IDF cosine similarity and count clusters where $\ge 3$ independent tool families converge on the same concern. Of the unique FP clusters, 41 are flagged by $\ge 3$ independent families --- roughly 30\% of the gold set size. Multi-tool agreement is suggestive but not conclusive: tools may share training biases that produce correlated false positives. Manual inspection of the top clusters confirms that many describe genuine bugs --- an off-by-one error in a CSS ordering mixin (10~tools), an unchecked \texttt{Optional.get()} risking \texttt{NoSuchElementException} (6~tools), unsanitized raw HTML enabling XSS (5~tools). Others are legitimate but low-severity findings that annotators may have deliberately excluded.

\paragraph{Issues the gold set arguably should not include.}
Six golden comments were caught by zero of the 20~tracked tools. Four are low-severity style concerns that experienced reviewers would typically skip. At least three other golden comments are factually incorrect or debatable: one penalises a tool for using a Python~3.13 API that the project targets; another flags an intentionally unimplemented abstract method that the original PR author confirmed as correct design.

\paragraph{Implications.}
These consensus false positives and zero-catch golden comments require closer human examination. The boundary between actionable issues and style nits must be sharpened: until it is codified, the gold set will systematically penalise tools whose severity threshold differs from the annotators' implicit threshold.

\section{Changepoints}
\label{app:changepoints}
\paragraph{Detection and null expectation.}
We apply PELT to each tool's daily F$_1$ series, then retain candidate shifts with $|\Delta| \geq 3.5$\,pp and permutation $p < 0.05$. Under the shuffled-time null, this threshold has a nominal false-positive rate of 5\% per candidate. Because we do not apply a family-wise correction across candidates and tools, we interpret changepoints alongside independent evidence rather than as causal proof.

\paragraph{Per-tool changelog evidence.}
The five corroborated changepoints, with external evidence:

\begin{itemize}
    \item \textbf{CodeRabbit (Feb 26, $+$5.2\,pp):} Three back-to-back changelog entries on Feb 24--26 (improved context gathering, reduced false positives, new review-comment threading). The PELT-detected shift falls on the final release date.
    \item \textbf{Codex (Apr 7, $+$4.1\,pp):} A model-availability update on Apr 7 retired older SKUs and routed all reviews through an updated model. Shift date matches exactly.
    \item \textbf{Greptile (Mar 18, $+$3.8\,pp):} The v4 engine release, announced the same week, overhauled diff-context windowing.
    \item \textbf{Greptile (Apr 12, $+$4.5\,pp):} A security-detection beta released Apr 11 added new rule categories; the shift straddles the two-day rollout.
    \item \textbf{Qodo (Feb 28, $+$6.1\,pp):} Changelog entries confirm a major prompt and model update. An internal version marker in the PR assembly text independently confirms the transition: filtering to the new marker reproduces the post-changepoint mean F$_1$; filtering to the old marker reproduces the pre-changepoint mean.
\end{itemize}

The remaining three changepoints lack public changelog matches but may reflect A/B tests or gradual rollouts not announced externally.

\begin{figure}[h]
\centering
\includegraphics[width=\textwidth]{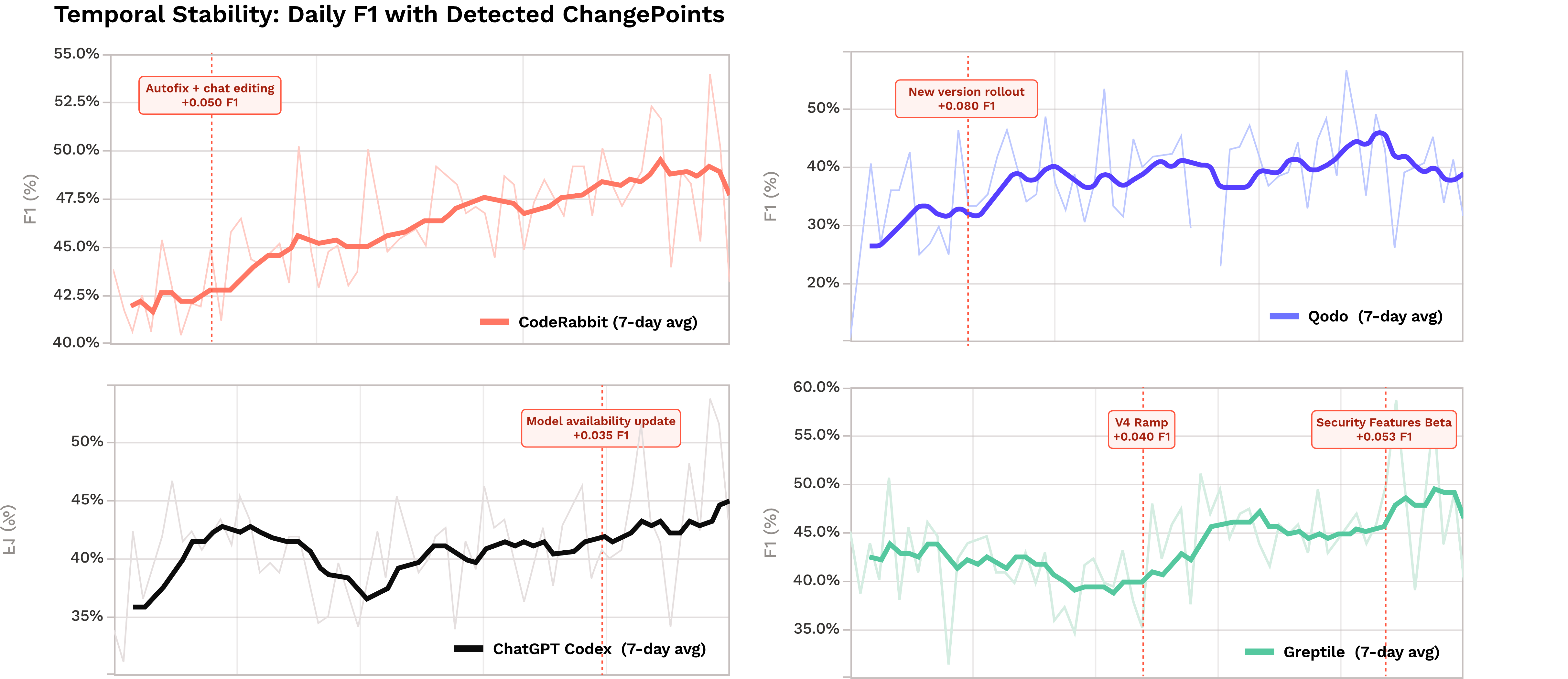}
\caption{Daily F$_1$ (thin line) and 7-day rolling average (thick line) for four tools with strong changelog-corroborated changepoints (Feb 16--Apr 21, 2026). Red dashed verticals mark level shifts detected by the PELT algorithm~\citep{killick2012} ($|\Delta| \geq 3.5$\,pp, permutation $p < 0.05$). White-boxed annotations show the matched public release event and the magnitude of the F$_1$ shift. Dotted gray horizontals indicate the pre- and post-changepoint mean F$_1$. A synchrony test across all tools yields $p_{\text{sync}} \approx 1.0$, so the detected shifts show no broad synchrony and are less consistent with a benchmark-wide judge or pipeline change. Of the eight detected changepoints across all tools, five align with independently verifiable vendor changelogs.}
\label{fig:changepoints}
\end{figure}

\label{app:tracking-changes}
\begin{figure}[h]
\centering
\includegraphics[width=\textwidth]{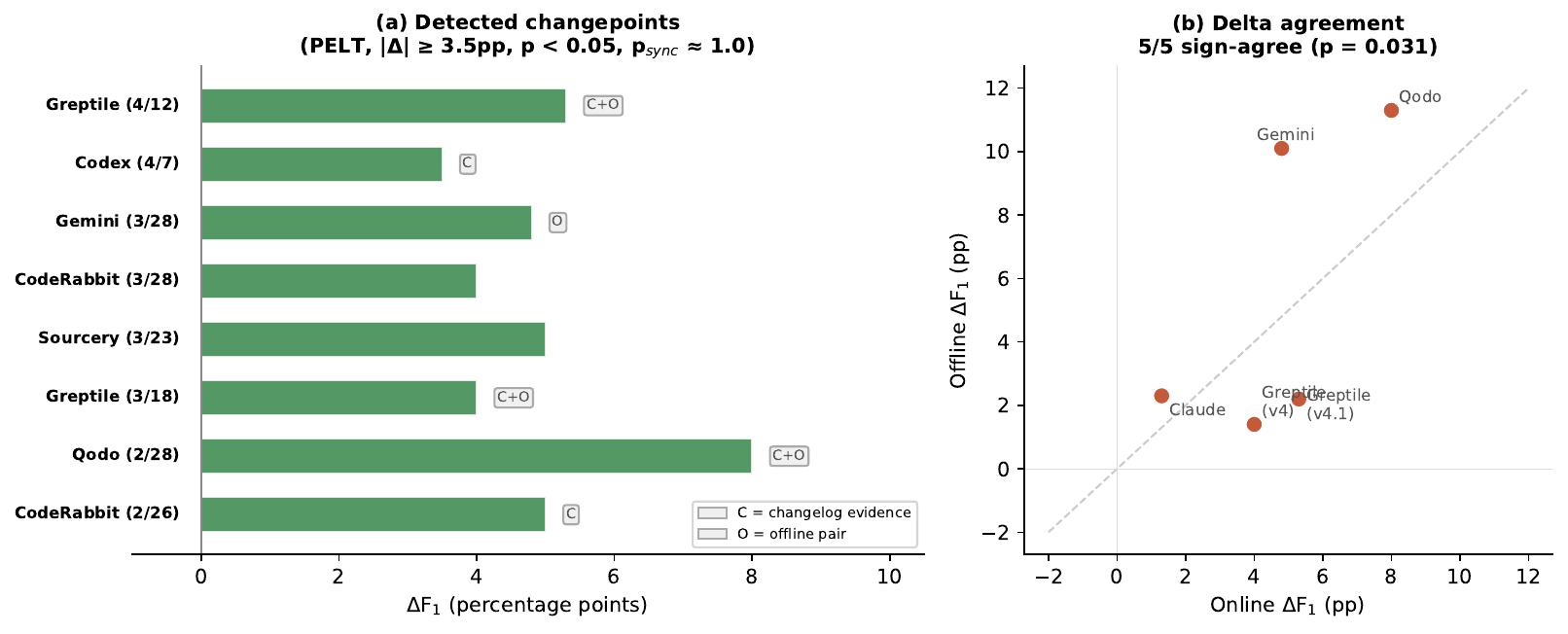}
\caption{Cross-benchmark agreement on tool-level changes (Feb 16--Apr 21, 2026).
\textbf{(a)}~Eight statistically significant changepoints detected by PELT ($|\Delta| \geq 3.5$\,pp, permutation $p < 0.05$), with external-evidence markers: C~=~public changelog entry, O~=~paired offline benchmark run. The synchrony statistic $p_{\text{sync}} \approx 1.0$ indicates that shifts are \emph{less} clustered than chance, providing evidence against a benchmark-wide artifact (a low $p_{\text{sync}}$ would instead suggest that many tools shifted together).
\textbf{(b)}~Online vs.\ offline $\Delta$F$_1$ for the 5 tool updates that have paired offline before/after runs. All five agree on sign (sign-test $p = 0.031$); four agree on magnitude within 3\,pp. This concordance provides the strongest evidence that the two benchmarks track the same underlying capability changes despite using different ground truths.}
\label{fig:tracking-changes}
\end{figure}

\section{Minimal Review Agent (MRA) results}
\label{app:mra}

To isolate the contribution of the product harness from the base model, we built a Minimal Review Agent (MRA): a fixed architecture of vector-store retrieval, tree-sitter context extraction, linting, and 11~specialized sub-agents. We ran it on the offline benchmark with two model stacks, holding the harness constant (See Table~\ref{tab:mra}).

\begin{table}[h]
\centering
\caption{Offline benchmark scores (Opus 4.5 as LLM judge) for three MRA configurations compared to commercial tools on the same 50-PR set. All MRA configs achieve identical recall despite using different model families and vastly different compute budgets, indicating the harness determines the score ceiling.}
\label{tab:mra}
\small
\begin{tabular}{lccccr}
\toprule
\textbf{System} & \textbf{P (\%)} & \textbf{R (\%)} & \textbf{F$_1$ (\%)} & \textbf{Tok/PR} & \textbf{Cost} \\
\midrule
\multicolumn{6}{l}{\emph{Commercial tools ($n=30$, 50~PRs each)}} \\
\quad Median       & 44.5 & 46.7 & 44.0 & --- & --- \\
\quad Mean         & 43.4 & 46.9 & 42.2 & --- & --- \\
\quad IQR          & [33.7, 48.2] & [38.1, 55.3] & [37.1, 46.5] & --- & --- \\
\midrule
\multicolumn{6}{l}{\emph{MRA configurations (same harness, different models, 50~PRs each)}} \\
mra-openai (GPT-5.5 + 5.4-mini)   & 16.6 & 35.0 & 22.5 & 1.0M & \$87 \\
mra-mix (GPT-5.5 + Haiku~4.5)     & 16.6 & 35.0 & 22.5 & 7.4M & \$491 \\
mra-claude (Sonnet~4.6 + Haiku~4.5) & 15.7 & 35.0 & 21.7 & 6.2M & \$408 \\
\bottomrule
\end{tabular}
\end{table}

\noindent\textbf{Key observations:}
\begin{itemize}
  \item All three configurations find the same total number of golden issues (48 true positives each out of 137), but they catch different \emph{subsets}: per-PR TP counts differ on 34 of 50~PRs across the three configs. Each model family identifies bugs the others miss, yet the harness caps total recall at the same level.
  \item Comparing mra-mix to mra-claude isolates the orchestrator effect (both use Haiku~4.5 subs): GPT-5.5 vs Sonnet~4.6 as orchestrator yields only a 1\,pp precision difference, confirming the orchestrator model has minimal impact.
  \item Comparing mra-mix to mra-openai isolates the sub-agent effect (both use GPT-5.5 orchestrator): swapping GPT-5.4-mini subs for Haiku~4.5 subs produces identical scores, confirming the sub-agent provider also does not matter.
  \item mra-claude and mra-mix consumed ${\sim}6$--$7{\times}$ more tokens per PR than mra-openai (6.2--7.4M vs 1.0M avg), largely due to thinking tokens and more tool calls per agent (${\sim}$26 vs ${\sim}$11). Despite $5$--$6{\times}$ higher cost, neither achieves better scores.
  \item All MRA configs fall well below commercial-tool precision (IQR 33.7--48.2\%), confirming that retrieval, filtering, and post-processing account for the gap, rather than the base model.
\end{itemize}

\paragraph{MIRA contrast.}
MIRA is a streamlined single-pass system that fetches the PR diff through the GitHub API and sends the diff and surrounding context to the model in one structured-output call, followed by a noise filter. It uses no sub-agents or agentic retrieval, and its optional repository index was disabled in our runs. Across four model configurations on the same benchmark, MIRA's recall spans 15.4\,pp and precision spans 29.8\,pp. This contrast shows that the flat aggregate recall observed in the three MRA runs is not universal: model choice has a much larger observed effect under MIRA's single-pass design.

\section{Review speed vs.\ quality}
\label{app:speed-quality}

We measure review latency as wall-clock time from trigger to last comment on the 50 offline benchmark PRs, and correlate median latency per tool with offline F$_1$ (Opus judge) across 25~tools. Median latencies range from $\sim$1 to $\sim$26 minutes. Neither Spearman ($\rho = 0.38$, $p = 0.06$) nor Pearson ($r = 0.19$, $p = 0.38$) correlation is significant, indicating that spending more wall-clock time does not reliably improve review quality on this benchmark.

%% --- Prompts ---

\section{Prompt templates}
\label{app:prompts}

The online judge uses a three-step structured-output pipeline. Each step receives the raw PR artifacts and produces JSON output consumed by the next step.

\subsection{Step 1: Extract bot suggestions}

\begin{small}
\begin{verbatim}
You are analyzing a pull request to extract all actionable
suggestions made by a code review bot.

The bot's username is: {bot_username}

Below you will see:
1. The commits that were under review (the code state the bot
   saw), including full diffs
2. The bot's review comments on those commits

For each actionable suggestion the bot made, extract:
- A unique ID (S1, S2, ...)
- A description of what was suggested
- The category (bug, style, performance, security, refactor,
  documentation, other)
- The file path and line number if available
- Severity (low, medium, high, critical)

Only include ACTIONABLE suggestions -- skip generic praise,
summaries, or "looks good" comments.
Skip bot comments that are purely informational without
suggesting any change.
\end{verbatim}
\end{small}

\subsection{Step 2: Extract developer actions}

\begin{small}
\begin{verbatim}
You are analyzing post-review commit diffs to extract every
concrete code issue that was fixed or improved AFTER the bot
reviewed the PR.

These extracted issues are ground-truth labels -- the actual
problems that existed in the code the bot reviewed. Your job
is to read the diffs and identify each distinct issue that
was addressed.

For each distinct code issue fixed in the post-review commits,
extract:
- A unique ID (A1, A2, ...)
- A description of the specific issue that was fixed (what was
  wrong, not what was done)
- The category (bug, style, performance, security, refactor,
  documentation, other)
- The file path where the fix was applied
- The commit SHA that fixed it
- Action type: why was this changed? (fix, improvement,
  cleanup, new_feature, other)

Guidelines:
- Focus on the DIFFS -- each action should correspond to a
  real code change you can see
- Describe the ISSUE that existed, not the PR activity
- One action per distinct issue
- Skip trivial formatting-only changes unless they fix a
  real problem
- Skip merge commits or changes that don't address a code issue
\end{verbatim}
\end{small}

\subsection{Step 3: Match suggestions to actions}

\begin{small}
\begin{verbatim}
You are judging whether a bot's code review suggestions
correspond to actual code issues that were later fixed.

You have two lists:
- Bot Suggestions: issues the bot flagged during review
- Code Fixes: actual issues that were fixed in post-review
  commits (ground truth)

For EACH bot suggestion, determine:
1. Does it match any code fix? (matched: true/false)
2. Which code fix? (human_action_id)
3. How confident are you? (0.0-1.0)
4. Brief reasoning

A suggestion is "matched" if:
- It identified the same issue (or substantially the same
  concern) that was later fixed
- The fix is in the same file/area the suggestion pointed to
- Even a partial overlap counts if the bot caught part of the
  real problem

A suggestion is NOT matched if:
- No corresponding fix exists
- The fix addresses a different concern than what the bot
  suggested
- The bot's suggestion was about something that wasn't a
  real problem
\end{verbatim}
\end{small}

\subsection{CAR judge (validation)}

The Comment Action Rate judge used for methodology robustness validation (\S\ref{app:judge-validation}). Unlike the matching judge which operates on extracted abstractions, the CAR judge operates directly on raw bot comments and post-review diffs.

\begin{small}
\begin{verbatim}
You are evaluating whether a bot's code review comment was
directly addressed by subsequent code changes.

You will see:
1. A bot review comment
2. The comment thread (replies and discussion)
3. Post-review code diffs

For the bot's comment, determine:
1. Was the concern directly addressed? (addressed: true/false)

   A comment is "addressed" ONLY if ALL of the following
   are true:
   - The code change is in the SAME FILE or closely related
     area that the comment references
   - The change is a DIRECT response to the specific concern
     raised -- not a coincidental change in the same area
   - There is clear evidence of causation: the developer
     either acknowledged the comment (e.g. "fixed", "done")
     OR the code change precisely implements what the bot
     suggested

   A comment is NOT addressed if:
   - The code change is in a different file/area
   - The overlap is marginal or coincidental
   - There is no acknowledgment AND the code change only
     loosely relates to the concern
   - The comment is a summary, observation, or praise --
     not an actionable suggestion
   - The comment describes what already exists rather than
     suggesting a change

2. How confident are you? (confidence: 0.0-1.0)
3. Brief reasoning explaining your judgment

IMPORTANT: Be strict. Only count clear, direct responses to
the bot's specific concern. Do not count indirect or
coincidental code changes.
\end{verbatim}
\end{small}

\section{Fine-tuning details}
\label{app:finetuning}

The fine-tuning experiment in §\ref{app:finetuning} is a minimal probe rather than a full-scale study. This appendix walks through the experimental constraints, the specific design choices that follow from them, and the results in detail.

\paragraph{Why the experiment is minimal.} Several constraints push against a full-scale fine-tuning study at the boundary of what open-weight tooling currently supports.

\emph{Context length.} Open-weight code models in the 7B class top out at 32K context, of which $\sim$28K is usable for prompt + response after the chat template. Real PRs in the corpus routinely exceed this --- at the 95th percentile, diffs run several thousand lines, well past the window even before retrieval. Any fine-tuning experiment at this scale therefore has to choose between truncating long PRs, dropping them, or restricting the training unit to something smaller than a PR. We do both: PR-level (full diff, with dropping above the limit) and hunk-level (target hunk plus surrounding context). The contrast between them is informative on its own.

\emph{PR size as the dominant difficulty.} The size-degradation finding in §\ref{sec:findings} --- F$_1$ drops 28 pp from small to extra-large PRs across the field --- means the corpus is most valuable exactly where current backbones cannot fit it. Fine-tuning on the easy half of the dataset is not the right experiment for closing the field's hardest gap.

\emph{Cost of full fine-tuning.} Full fine-tuning a 7B model end-to-end is expensive enough that thorough hyperparameter sweeps, multi-seed averaging, and ablations across base models are out of scope for a probing experiment. We use LoRA adapters (rank 8, $\alpha=16$, target modules \texttt{q\_proj}/\texttt{v\_proj}; $\sim$2.5M trainable parameters, 0.03\% of total), single seed, a single A100 40GB. This is the lightest setup that can still run DPO end-to-end at 32K context.

\emph{The competitive landscape.} Frontier commercial review tools combine years of product iteration with retrieval, linting, multi-agent orchestration, and prompt engineering on top of much larger base models. A single LoRA fine-tuning run on a 7B base is not going to outperform them, and was not designed to. The MRA experiment in §\ref{sec:findings} already shows the harness gap dwarfs the model gap; this experiment tests the orthogonal question of whether the corpus carries learnable signal at all.

\paragraph{What the experiment can and cannot answer.} Within these constraints, the experiment is structured around a single clean question: does flipping the labels degrade the model? If yes, the labels carry direction-bearing signal a model can learn from. The questions this experiment cannot answer --- whether scaled-up fine-tuning closes the gap to commercial tools, what the optimal recipe looks like, how the signal interacts with longer context, whether the gains generalize across base models --- require resources outside the scope of any single paper. We release the dataset to make those experiments possible.

\paragraph{Setup.} Both pipelines share Qwen2.5-Coder-7B-Instruct with LoRA adapters as above. Stage 1 is supervised fine-tuning on accepted suggestions only --- the model learns what a human-actionable review comment looks like. Stage 2 is direct preference optimization on (accepted, ignored) pairs constructed within the same PR and the same bot, length-matched to within 20\% to prevent the model from learning length cues, and bot-balanced so no single tool dominates. Held-out by repository, with PRs after 2026-03-31 reserved for evaluation. CodeRabbit is excluded for compliance reasons unrelated to the experiment.

\paragraph{Ablation arms.} Three arms vary the DPO label assignment while holding the training recipe constant: \texttt{filtered} (real labels with the §\ref{sec:noise} quality filters) and \texttt{inverted} (\texttt{chosen}/\texttt{rejected} swapped within each pair). \texttt{filtered} vs \texttt{inverted} is the central contrast: same data structure, same compute, only the direction of the labels differs.

\paragraph{Evaluation.} 60 held-out PRs the model never trained against, $k=3$ completions per prompt at temperature 0.7, matched to gold human actions with the same matcher used in §\ref{sec:methodology}. F$_1$ reported with 95\% bootstrap CIs.

\paragraph{Results.} Table~\ref{tab:finetuning} reports F$_1$ per arm. The headline is the cross-pipeline replication of the filtered-vs-inverted contrast: both pipelines show filtered above inverted by similar magnitudes despite differing in the training unit. PR-level filtered beats base by 3.3 pp; hunk-level fine-tuning underperforms base overall, with SFT alone regressing furthest. We read the hunk-level base regression as a constraint of the truncated training unit rather than a failure of the labels: the within-DPO contrast holds on hunk-level anyway, indicating the labels carry signal that survives context truncation even when the absolute fine-tuning recipe does not.

\begin{table}[h]
\centering
\small
\begin{tabular}{lcc}
\toprule
\textbf{Arm} & \textbf{PR-level F$_1$} & \textbf{Hunk-level F$_1$} \\
\midrule
Base (untuned Qwen) & 0.530 [0.41, 0.64] & 0.526 [0.37, 0.69] \\
SFT-only            & 0.562 [0.47, 0.66] & 0.413 [0.29, 0.54] \\
DPO \texttt{filtered} & \textbf{0.563} [0.48, 0.66] & \textbf{0.495} [0.36, 0.63] \\
DPO \texttt{inverted} & 0.502 [0.40, 0.61] & 0.409 [0.31, 0.52] \\

\bottomrule
\end{tabular}
\caption{F$_1$ against gold human actions, 60 held-out PRs, $k=3$ samples per prompt, single seed, 95\% bootstrap CIs in brackets. The filtered-vs-inverted contrast holds on both pipelines (PR-level: 6.1 pp gap; hunk-level: 8.6 pp gap), evidence that the gain is label-driven rather than compute-driven.}
\label{tab:finetuning}
\end{table}

\paragraph{Limitations.} N=60 evaluation PRs, $k=3$ samples, single seed --- CIs are wide for everything, and the cross-pipeline agreement is doing more work than any single contrast. The \texttt{random} arm is queued; it adds a structural placebo that preserves pair counts and length distributions while destroying within-pair signal, which would further constrain what the filtered-vs-inverted contrast is sensitive to. Larger base models with longer effective context are likely to extract more signal from the same data, particularly on the size regime where commercial tools also degrade most (§\ref{sec:findings}); the dataset is sized to support that experiment but we do not run it here.

\section{Goodhart's Law: gaming vectors and countermeasures}
\label{app:goodhart}

The benchmark's primary defense against Goodhart's Law is structural: the reward signal is revealed developer behavior at scale, which is expensive to manipulate because it requires changing how thousands of independent developers respond to suggestions across diverse repositories. However, no benchmark is immune to gaming, and we identify several plausible attack vectors along with the countermeasures currently in place or planned. These vectors are not exhaustive; ongoing monitoring may reveal new strategies that require additional filters or detection mechanisms.

\paragraph{Vector 1: Selective deployment.}
A vendor could inflate scores by deploying only in repositories with high acceptance rates (e.g., solo-developer projects where the author rubber-stamps all suggestions).

\textit{Countermeasure:} The concentration cap (Table~\ref{tab:filters}) limits any single author--repo--bot triple to $\leq$50 PRs, preventing score domination by favorable repositories. The ``team repos'' and ``human reviewer'' strict filters further restrict to multi-contributor projects with active human oversight. Rankings remain stable across filter tiers ($\rho \geq 0.78$; App.~\ref{app:engagement-sensitivity}), indicating this vector does not currently drive scores.

\paragraph{Vector 2: Judge alignment.}
A vendor could craft suggestions that are easy for the LLM judge to match to developer actions, even if the actual causal link is weak---effectively optimizing for the measurement instrument rather than actual developer utility.

\textit{Countermeasure:} Task~2 human validation (\S\ref{sec:human-validation}) provides an independent check: if a tool's automated precision is high but human experts rate its comments poorly, the discrepancy would be visible. Cross-judge agreement (Fleiss's $\kappa = 0.86$) and the alternative Comment Action Rate methodology (which skips extraction entirely) reaching near-identical scores suggest the judge is not easily fooled by surface-level optimization. As the benchmark matures, periodic human spot-checks provide an ongoing calibration signal.

\paragraph{Vector 3: Low-effort suggestion flooding.}
A vendor could generate many trivially-actionable suggestions (e.g., whitespace fixes, import ordering) to boost recall without providing substantive value.

\textit{Countermeasure:} The engagement filter requires visible developer response, which penalizes noise (developers learn to ignore tools that flood trivial suggestions). More directly, precision penalizes verbosity: flooding with low-value suggestions that developers ignore reduces precision proportionally. F$_1$'s harmonic mean ensures that recall gains from flooding are offset by precision losses. Task~2 provides a further check: annotators rate bot comments for actionability, not just accuracy, so a tool that scores high on recall but low on human-rated actionability would be flagged.

\paragraph{Vector 4: Competitor sabotage.}
An actor could create repositories with competing tools installed and systematically ignore their suggestions, tanking those tools' precision scores.

\textit{Countermeasure:} The concentration cap limits per-author--repo contribution, so a single actor's repositories have bounded influence. The ``team repos'' filter (requiring $\geq$2 contributors) excludes solo-operated repositories entirely. More fundamentally, this attack scales poorly: each tool's score is aggregated across thousands of PRs from hundreds of independent repositories, so a meaningful score shift would require coordinated non-engagement across many accounts and repositories, which is both expensive and detectable via anomaly monitoring (e.g., clusters of repos with unusually low engagement for a single tool).

\paragraph{Vector 5: Temporal manipulation.}
A vendor could temporarily improve their tool during known measurement windows and revert afterward.

\textit{Countermeasure:} The benchmark operates continuously with rolling averages, not fixed evaluation windows. The PELT changepoint detector (\S\ref{sec:findings}) identifies statistically significant shifts in tool performance; a tool that oscillates would produce frequent changepoints, which would be visible to any observer of the public dashboard.

\paragraph{Residual risk.}
The most sophisticated gaming strategy would involve genuinely improving the tool to produce suggestions that developers act on---but this is indistinguishable from legitimate improvement, which is exactly the behavior the benchmark intends to incentivize. The remaining gap is between ``developer acts on it'' and ``it was actually valuable in context,'' which Task~2 bounds at $\rho = 0.688$ correlation. We view this as the fundamental irreducible limitation of any behavioral metric, and recommend that benchmark consumers use scores alongside qualitative evaluation rather than as a sole decision criterion.

%% file: tables/online-offline-contrasts.tex
\begin{table}[h]
\centering
\small
\caption{How the online and offline benchmarks differ and complement each other.}
\label{tab:contrasts}
\begin{tabular}{p{2cm} p{4.9cm} p{4.9cm}}
\toprule
\textbf{Dimension} & \textbf{Online} & \textbf{Offline} \\
\midrule
Ground truth & Revealed developer actions (PR author decides) & Expert-curated gold set (third-party annotator decides) \\
\addlinespace
Scale & 500K+ PRs, refreshed daily & 50 PRs, fixed set \\
\addlinespace
Head-to-head & Not possible (one tool per repo) & All tools review the same PRs \\
\addlinespace
Temporal & Continuously updated; detects changepoints & Point-in-time snapshot; confirms direction of change \\
\addlinespace
P/R regime & Precision-leaning (mean P/R = 1.53): developers engage selectively & Recall-leaning (mean P/R = 0.80): comprehensive gold set with known issues \\
\addlinespace
Tool config & As deployed (org-specific knobs, severity thresholds) & Default config (raw capability) \\
\addlinespace
Annotator alignment & Misaligned with bot: developer is author, bot is third party & Aligned with bot: both annotator and bot are third-party reviewers \\
\addlinespace
Goodhart risk & Low (new data daily, requires changing real behavior at scale) & Higher (fixed set can leak into training) \\
\bottomrule
\end{tabular}
\end{table}

%% file: tables/beta-sensitivity.tex
\begin{table}[h]
\centering
\caption{Tool rankings under F$_{0.5}$, F$_1$, and F$_2$.
  $\Delta$rank shows the rank change from F$_{0.5}$ to F$_2$; negative values indicate the tool benefits when recall is weighted more. 
  }
\label{tab:beta-sensitivity}
\small
\begin{tabular}{l c c c c c c}
\toprule
Tool & P & R & F$_{0.5}$ (rank) & F$_1$ (rank) & F$_2$ (rank) & $\Delta$rank \\
\midrule
Cubic      & .761 & .469 & .674 (1)  & .580 (1)  & .507 (2)  & +1 \\
CodeRabbit & .554 & .499 & .542 (8)  & .525 (2)  & .510 (1)  & $-$7 \\
Gemini     & .651 & .438 & .594 (3)  & .524 (3)  & .469 (5)  & +2 \\
Sourcery   & .583 & .470 & .559 (7)  & .520 (4)  & .492 (3)  & $-$4 \\
Claude     & .591 & .458 & .560 (6)  & .516 (5)  & .481 (4)  & $-$2 \\
Greptile   & .728 & .374 & .613 (2)  & .494 (6)  & .414 (7)  & +5 \\
Qodo       & .565 & .389 & .517 (11) & .461 (7)  & .421 (6)  & $-$5 \\
Cursor     & .685 & .345 & .574 (4)  & .459 (8)  & .383 (8)  & +4 \\
Codex      & .674 & .337 & .561 (5)  & .450 (9)  & .375 (9)  & +4 \\
Copilot    & .609 & .338 & .525 (10) & .434 (10) & .370 (10) & 0 \\
Devin      & .647 & .302 & .526 (9)  & .412 (11) & .338 (11) & +2 \\
\bottomrule
\end{tabular}
\end{table}

%% file: tables/bot-coverage.tex
\begin{table}[t]
\centering
\caption{Per-tool coverage in the enriched dataset.
  \textbf{PRs}: total enriched pull requests.
  \textbf{Repos}: distinct repositories.
  \textbf{PR/Repo}: concentration ratio.
  \textbf{Contributor columns}: fraction of a tool's repositories
  whose PRs come from exactly 1, 2, 3--5, 6--10, or 11+ distinct authors.}
\label{tab:bot-coverage}
\small
\begin{tabular}{l r r r r r r r r}
\toprule
    Tool & PRs & Repos & PR/Repo & 1 & 2 & 3--5 & 6--10 & 11+ \\
\midrule
    Copilot       & 142{,}779 & 70{,}000 &  2.0 & 74.2 &  5.9 &  1.4 & 0.2 & 0.1 \\
    CodeRabbit    & 130{,}980 & 33{,}925 &  3.9 & 77.0 &  9.9 &  5.2 & 1.3 & 0.8 \\
    Gemini        & 114{,}212 & 18{,}107 &  6.3 & 75.5 &  8.2 &  4.1 & 1.2 & 0.8 \\
    Codex         & 113{,}232 & 36{,}251 &  3.1 & 81.9 &  4.1 &  0.8 & 0.2 & 0.1 \\
    Greptile      &  79{,}243 &  4{,}870 & 16.3 & 65.1 & 16.8 &  8.6 & 2.3 & 2.7 \\
    Cursor        &  69{,}906 & 19{,}512 &  3.6 & 86.1 &  4.9 &  2.1 & 0.6 & 0.4 \\
    Claude        &  62{,}710 &  6{,}363 &  9.9 & 77.8 & 11.5 &  5.1 & 1.5 & 0.8 \\
    Devin         &  53{,}976 &  7{,}526 &  7.2 & 72.6 &  7.2 &  2.0 & 0.4 & 0.4 \\
    Sourcery      &  40{,}154 &  6{,}587 &  6.1 & 76.4 &  9.6 &  3.8 & 1.0 & 0.7 \\
    Qodo          &  30{,}802 &  5{,}343 &  5.8 & 81.7 & 11.0 &  3.9 & 0.5 & 0.5 \\
    Cubic         &  24{,}971 &  2{,}842 &  8.8 & 63.1 & 12.0 &  5.5 & 1.4 & 1.3 \\
    CodeAnt       &   9{,}413 &  1{,}516 &  6.2 & 82.5 & 11.5 &  2.2 & 0.3 & 0.2 \\
    Graphite      &   8{,}588 &    399   & 21.5 & 59.6 & 10.8 & 12.3 & 6.3 & 6.0 \\
    Augment       &   6{,}643 &    946   &  7.0 & 77.0 & 10.9 &  3.7 & 1.0 & 0.3 \\
    Baz           &   1{,}656 &    106   & 15.6 & 72.6 & 12.3 &  6.6 & 0.9 & 1.9 \\
    Kody          &   1{,}304 &    168   &  7.8 & 76.2 & 14.3 &  6.5 & 1.8 & 0.0 \\
    Propel        &   1{,}118 &    136   &  8.2 & 77.2 &  5.1 &  1.5 & 2.2 & 2.2 \\
    Mesa          &   1{,}087 &    132   &  8.2 & 75.0 & 22.7 &  0.0 & 0.0 & 0.0 \\
    \midrule
    \textbf{Total} & \textbf{892{,}774} & \textbf{189{,}837} & & & & & & \\
    \bottomrule
    \end{tabular}
\end{table}

%% file: tables/validation-inventory.tex
\begin{table}[h]
\centering
\label{tab:validation-inventory}
\small
\caption{This table illustrates internal validation tests. Each row reports a single test, the method, and the result. Detailed methodology and figures are referenced in the rightmost column.}
\begin{tabular}{p{3.1cm} p{5.0cm} p{3.3cm} p{.9cm}}
\toprule
\textbf{Test} & \textbf{Method} & \textbf{Result} & \textbf{Ref.} \\
\midrule
\multicolumn{4}{l}{\textit{Instrument robustness}} \\
\midrule
Cross-judge agreement (matching) & Three judges, shared extractions & Fleiss's $\kappa = 0.86$ & App.~\ref{app:judge-validation} \\
Cross-judge agreement (full pipeline) & Three judges, independent runs & $\rho = 0.69$--$0.81$ pairwise & App.~\ref{app:judge-validation} \\
Alternative judge (CAR) & Skip extraction; match comments to diffs directly & 82\% binary agreement; mean bias $<0.01$ & App.~\ref{app:judge-validation} \\
Filter convergence & Re-rank under six filter configurations & $\rho \geq 0.78$ pairwise & App.~\ref{app:filter-convergence} \\
Filter sensitivity & Sweep contributor/reviewer thresholds & $<2$pp shift in precision; $<5$pp in recall & App.~\ref{app:engagement-sensitivity} \\
\midrule
\multicolumn{4}{l}{\textit{Replicability}} \\
\midrule
Chronological split-half & Re-rank on each half of 64-day window & $\rho = 0.80$ & App.~\ref{app:temporal-stability} \\
Random split-half & 1{,}000 random day-partitions & Median $\rho = 0.88$, 95\% CI $[0.66, 0.97]$ & App.~\ref{app:temporal-stability} \\
Window-size stability & 14-day blocks, thirds, halves & All $\rho > 0.79$ & App.~\ref{app:temporal-stability} \\
Bootstrap noise floor & Day-block bootstrap, 2{,}000 resamples & Median SE $= 0.005$; 61\% of pairs distinguishable & App.~\ref{app:noise-floor} \\
\midrule
\multicolumn{4}{l}{\textit{Fairness}} \\
\midrule
DML confound robustness & Control for 9 observable confounders & $\rho = 0.86$ raw vs.\ debiased & App.~\ref{app:dml} \\
\bottomrule
\end{tabular}
\end{table}

%% file: tables/judge-agreement.tex
\begin{table}[h]
\centering
\caption{Inter-judge agreement on 180 stratified PRs. \emph{Matching only}: models share extractions and independently match suggestions to actions. \emph{Full pipeline}: each model runs extraction and matching independently.}
\label{tab:judge-agreement}
\small
\begin{tabular}{llr}
\toprule
\textbf{Comparison} & \textbf{Metric} & \textbf{Value} \\
\midrule
\multicolumn{3}{l}{\emph{Matching only (shared extractions)}} \\
All three models        & Fleiss's $\kappa$    & 0.86 \\
GPT-5 Nano vs.\ Claude Sonnet 4.5 & Cohen's $\kappa$ & 0.90 \\
GPT-5 Nano vs.\ Gemini 3 Flash    & Cohen's $\kappa$ & 0.84 \\
Claude Sonnet 4.5 vs.\ Gemini 3 Flash & Cohen's $\kappa$ & 0.82 \\
\midrule
\multicolumn{3}{l}{\emph{Full pipeline (independent extraction + matching)}} \\
GPT-5 Nano vs.\ Claude Sonnet 4.5 & Spearman $\rho$ & 0.81 \\
GPT-5 Nano vs.\ Gemini 3 Flash    & Spearman $\rho$ & 0.69 \\
Claude Sonnet 4.5 vs.\ Gemini 3 Flash & Spearman $\rho$ & 0.74 \\
\midrule
\multicolumn{3}{l}{\emph{Methodology robustness (Matching judge vs.\ CAR judge, $n = 172$)}} \\
Binary agreement        & \%           & 82 \\
Inter-methodology       & Cohen's $\kappa$ & 0.63 \\
Per-PR correlation      & Spearman $\rho$  & 0.59 \\
Mean score bias         & $\Delta$     & $<$0.01 \\
\bottomrule
\end{tabular}
\end{table}

%% file: tables/engagement-sensitivity.tex
\begin{table}[h]
\centering
\caption{Precision and recall under progressively stricter filter thresholds. All values are weighted averages across qualifying tools.}
\label{tab:engagement-sensitivity}
\small
\begin{tabular}{llrrr}
\toprule
\textbf{Dimension} & \textbf{Threshold} & \textbf{Precision} & \textbf{Recall} & \textbf{Scored PRs} \\
\midrule
\multirow{5}{*}{Repo contributors}
 & 1+ (default) & 0.596 & 0.415 & 84{,}183 \\
 & 2+           & 0.589 & 0.404 & 40{,}078 \\
 & 3+           & 0.586 & 0.404 & 26{,}843 \\
 & 6+           & 0.602 & 0.398 & 14{,}932 \\
 & 11+          & 0.605 & 0.400 &  9{,}044 \\
\midrule
\multirow{3}{*}{Human reviewers}
 & 0+ (default) & 0.596 & 0.415 & 88{,}194 \\
 & 1+           & 0.587 & 0.403 & 20{,}969 \\
 & 2+           & 0.551 & 0.373 &  3{,}931 \\
\midrule
\multirow{2}{*}{Engagement filter}
 & Required (default) & 0.596 & 0.415 & 88{,}155 \\
 & Not required       & 0.592 & 0.415 & 98{,}774 \\
\bottomrule
\end{tabular}
\end{table}

%% file: tables/dml.tex
\begin{table}[h]
\centering
\caption{DML debiased tool effects ($\hat{\theta}$) vs.\ raw differences. $^*$~denotes $p < 0.05$. See text above for column definitions.}
\label{tab:dml}
\small
\begin{tabular}{lrrrr}
\toprule
\textbf{Tool} & \textbf{$\hat{\theta}$} & \textbf{SE} & \textbf{Raw diff} & \textbf{$n$} \\
\midrule
Cubic Dev AI     & $+$0.117$^*$ & 0.008 & $+$0.158 & 2{,}695 \\
Devin            & $+$0.046$^*$ & 0.010 & $+$0.115 & 2{,}414 \\
Greptile         & $+$0.023$^*$ & 0.006 & $+$0.070 & 7{,}978 \\
Cursor           & $+$0.021$^*$ & 0.007 & $+$0.089 & 4{,}607 \\
Claude           & $+$0.012$^*$ & 0.005 & $+$0.001 & 6{,}123 \\
Copilot          & $+$0.012$^*$ & 0.005 & $-$0.015 & 10{,}308 \\
Codex            & $+$0.005     & 0.007 & $+$0.046 & 5{,}150 \\
Gemini           & $-$0.024$^*$ & 0.004 & $-$0.016 & 11{,}539 \\
Sourcery         & $-$0.024$^*$ & 0.007 & $-$0.070 & 3{,}902 \\
Qodo             & $-$0.025$^*$ & 0.009 & $-$0.065 & 1{,}747 \\
CodeRabbit       & $-$0.030$^*$ & 0.004 & $-$0.074 & 20{,}560 \\
CodeAnt          & $-$0.049$^*$ & 0.020 & $-$0.074 & 318 \\
\bottomrule
\end{tabular}
\end{table}